\documentclass{nature}

\usepackage{graphicx}
\usepackage{amsmath}
\usepackage{amsthm}
\usepackage{amssymb}
\usepackage{color}
\usepackage{algorithm}
\usepackage{algorithmic}
\usepackage{hyperref}

\usepackage[font=singlespacing]{caption}
\usepackage{setspace}
\usepackage{times}

\DeclareMathOperator{\E}{\mathbb{E}}

\newcommand{\ww}{v^{\text{wr}}}

\newcommand{\sg}[1]{\text{StopGradient}({#1})}

\DeclareMathOperator{\Tr}{Tr}

\let\Algorithm\algorithm
\renewcommand\algorithm[1][]{\Algorithm[#1]\setstretch{1.}}

\title{Optimizing Agent Behavior over Long Time Scales by Transporting Value} 

\author{\raggedright{Chia-Chun Hung$^{1\ast\dagger}$, Timothy Lillicrap$^{1\ast\dagger}$, Josh Abramson$^{1\ast}$, Yan Wu$^{1}$, Mehdi Mirza$^{1}$, Federico Carnevale$^{1}$, Arun Ahuja$^{1}$, Greg Wayne$^{1\ast\dagger}$}.\\
 $^1$DeepMind, 5 New Street Square, London EC4A 3TW, UK. \\
$^*$These authors contributed equally to this work. \\
$^\dagger$To whom correspondence should be addressed.}

\date{}

\begin{document} 

\singlespacing
\setlength\parindent{0pt}
\maketitle
\vspace{0.5cm}


\maketitle 

\begin{abstract}
Humans spend a remarkable fraction of waking life engaged in acts of ``mental time travel''\cite{corballis2014recursive}. We dwell on our actions in the past and experience satisfaction or regret. More than merely autobiographical storytelling, we use these event recollections to change how we will act in similar scenarios in the future. 
This process endows us with a computationally important ability to link actions and consequences across long spans of time, which
figures prominently in addressing the problem of \emph{long-term temporal credit assignment}; in artificial intelligence (AI) this is the question of how to evaluate the utility of the actions within a long-duration behavioral sequence leading to success or failure in a task. 
Existing approaches to shorter-term credit assignment in AI cannot solve tasks with long delays between actions and consequences. 
Here, we introduce a new paradigm for reinforcement learning where agents use recall of specific memories to credit actions from the past, allowing them to solve problems that are intractable for existing algorithms. 
This paradigm broadens the scope of problems that can be investigated in AI and offers a mechanistic account of behaviors that may inspire computational models in neuroscience, psychology, and behavioral economics.
\end{abstract}

The theory of how humans and animals express preferences and make decisions to ensure future welfare is a question of long-standing concern, dating to the origins of economic utility theory\cite{samuelson1937note}. Within multiple fields, including economics and behavioral psychology, there remains unresolved debate about the appropriate formalism to explain valuation of temporally distant reward outcomes in long-term decision making. 

In AI research, the problem of how to learn rational behavior that is temporally far-sighted is known as the credit assignment problem\cite{newell1955chess, samuel1959some, minsky1961steps}. An AI agent must evaluate the utility of individual actions within a long sequence. To address the credit assignment problem, deep learning has been combined with reinforcement learning (RL) to provide a flexible class of architectures and algorithms that can be used practically to estimate the utility of courses of action for behaving agent models engaged in sensorimotor tasks in complex environments.

These algorithms have almost exclusively borrowed the assumptions of discounted utility theory\cite{samuelson1937note, thomas2014bias, sutton1998reinforcement} and achieve credit assignment using value function bootstrapping and backpropagation\cite{mnih2015human} (deep RL). Practical and convergent deep RL algorithms discount the future, reducing their applicability for problems with long delays between decisions and consequences\cite{baxter2001infinite, schulman2016optimizing}.

Conspicuously, humans and animals evidence behaviors that deep RL cannot yet simulate behaviorally. These come for example under the headings of latent learning\cite{blodgett1929effect, tolman1948cognitive}, prospective memory\cite{mcdaniel2004delaying}, and inter-temporal choice\cite{frederick2002time}, and encompass learning and decision-making that occurs either without task reward, or when rewards are recouped at long delay from relevant choice points. It has been argued that hominid cognitive ability became truly modern when new strategies for long-term temporal credit assignment and planning emerged, leading to abrupt cultural shifts and immense changes in social complexity and human achievement\cite{klein2002dawn}. Algorithmic progress on problems of long-term credit assignment may similarly lead to profound magnification of the range of decision-making problems that can be addressed computationally.

Our paradigm builds on deep RL but introduces a new set of principles for credit assignment over long time scales, the problem of long-term temporal credit assignment. First, agents must encode and store perceptual and event memories; second, agents must predict future rewards by identifying and accessing memories of those past events; third, they must revaluate these past events based on their contribution to future reward. 

Based on these principles we introduce a new algorithm, called Temporal Value Transport (TVT), which uses neural network attentional memory mechanisms to credit distant past actions for future rewards. This algorithm automatically splices together temporally discontiguous events, identified by task relevance and their association to each other, allowing agents to link actions with their ultimate consequences. The algorithm is not without heuristic elements, but we prove its effectiveness for a set of tasks requiring long-term temporal credit assignment over delay periods that pose enormous difficulties to conventional deep RL.

We formally consider the widely used setting of \emph{episodic reinforcement learning} (episodic RL), where time is divided into separate trials or \emph{episodes}, with a distribution of starting states, and terminating after $T$ time steps. The agent's behavior is governed by a set of tuneable parameters $\theta$, and it operates in the environment by receiving at each discrete time step $t$ sensory observations $o_t$, processing those observations into an internal representation $h_t = h(o_0, \dots, o_t; \theta)$, and emitting actions $a_t$ using a ``policy'' probability distribution $\pi(a_t | h_t, y_t; \theta)$ ($y_t$ is included to allow for conditioning variables, which will be used later). Each episode is independent of the rest save for any changes due to learning of the agent itself.

The objective of episodic RL is to maximize the sum of rewards that the agent receives until the final time step. Let $\mathcal{R}_t \equiv r_t + r_{t+1} + r_{t+2} + \dots + r_T$, where $r_t$ is the reward at time step $t$ and $\mathcal{R}_t$ is called the return. The return of any episode is non-deterministic due to randomness in the start state of the system and the random action choices of the policy. Therefore, beginning from the start of the episode the aim is to maximize the expected return, known as the \emph{value}  
\begin{align}
\label{eq:episodic}
V_0 & = \E_{\pi} [\mathcal{R}_0] \nonumber \\
& = \E_{\pi} \bigg[\sum_{t=0}^T r_t \bigg].
\end{align}
To improve performance, it is common to evaluate the \emph{episodic policy gradient}\cite{williams1992simple, sutton2000policy}, which under fairly general conditions can be shown to have the form:
\begin{align}
\label{eq:episodic_pg}
\nabla_\theta V_0 & = \nabla_\theta \E_{\pi} \bigg[\sum_{t=0}^T r_t \bigg] \nonumber \\
& = \E_{\pi} \bigg[\sum_{t=0}^T \nabla_{\theta} \log \pi(a_t | h_t; \theta) \mathcal{R}_t \bigg],
\end{align}
where $\nabla_\theta$ is the gradient with respect to $\theta$.
This quantity is typically estimated by running episodes and sampling actions from the probability distribution defined by the policy and calculating at each episode:
\begin{align}
\label{eq:episodic_pg_mc}
\nabla_\theta V_0 & \approx \Delta \theta = \sum_{t=0}^T \nabla_{\theta} \log \pi(a_t | h_t; \theta) \mathcal{R}_t.  
\end{align}
In practice, updating the parameters of the agent using Eq.~\ref{eq:episodic_pg_mc} is only appropriate for the simplest of reinforcement learning tasks because, though its expectation is the episodic policy gradient, it is a stochastic estimate with high variance. That is, for the gradient estimate $\Delta \theta$, $\text{Var}_\pi(\Delta \theta)$ is large relative to the magnitude of the expectation in Eq.~\ref{eq:episodic_pg}. Most practical applications of reinforcement learning mitigate this variance in two ways. First, they utilize variance reduction techniques, including, for example, replacing $\mathcal{R}_t$ by a mean-subtracted / ``baselined'' estimate $\mathcal{R}_t - \hat{V}_t$, where $\hat{V}_t$ is a learned prediction of $\mathcal{R}_t$\cite{sutton1998reinforcement}. In this work, we use variance reduction techniques, but we will sometimes suppress mention of them in the primary exposition when they are not our focus (Supplement Section 2.2).

Another approach to reducing variance is to introduce \emph{statistical bias}\cite{geman1992neural}: i.e., by choosing a direction of update to the parameters $\Delta \theta$ that does not satisfy $\E_\pi[\Delta \theta] = \nabla_\theta V_0$. 
One of the most common tools used to manipulate bias to reduce variance is temporal discounting, which diminishes the effect of future rewards on the gradient. We define the discounted return as $\mathcal{R}_t^{(\gamma)} = r_t + \gamma r_{t+1} + \gamma^2 r_{t+2} + \dots + \gamma^{T-t} r_T$. The parameter $\gamma \in [0,1]$ is known as the \emph{discount factor} (cf.~discount rate in economics\cite{frederick2002time}). For $\gamma = 0.99$, a reward $100 \, (= \frac{1}{1-\gamma})$ steps into the future is attenuated by a multiplicative factor of
\begin{align}
0.99^{100} =
\bigg(1 - \frac{1}{100}\bigg)^{100} \approx 1/e.    
\end{align}
In general, the \emph{half-life} (strictly, the $1/e$-life) of reward in units of time steps is $\tau = \frac{1}{1-\gamma}$. Because effectively fewer reward terms are included in the policy gradient, the variance of the discounted policy gradient estimate
\begin{align}
\label{eq:episodic_pg_mc_discounted}
\nabla_\theta V_0^{(\gamma)} & \approx \sum_{t=0}^T \nabla_{\theta} \log \pi(a_t | h_t; \theta) \mathcal{R}_t^{(\gamma)}  
\end{align}
is smaller. Unfortunately, because the influence of future reward on present value is exponentially diminished, discounting limits the largest time scale to which an agent's behavior is adapted to roughly a multiple of the half-life. Due to this limitation, RL research and applications focus on relatively short time-scale problems such as reactive video games\cite{mnih2015human}. Yet clearly there is a gap between these tractable time scales and relevant human time scales: much of the ``narrative structure'' of human life is characterized by highly-correlated, sparse events that are separated by long time intervals and unrelated, intervening activities.

\begin{figure}
    \centering
    \includegraphics[width=0.99\textwidth]{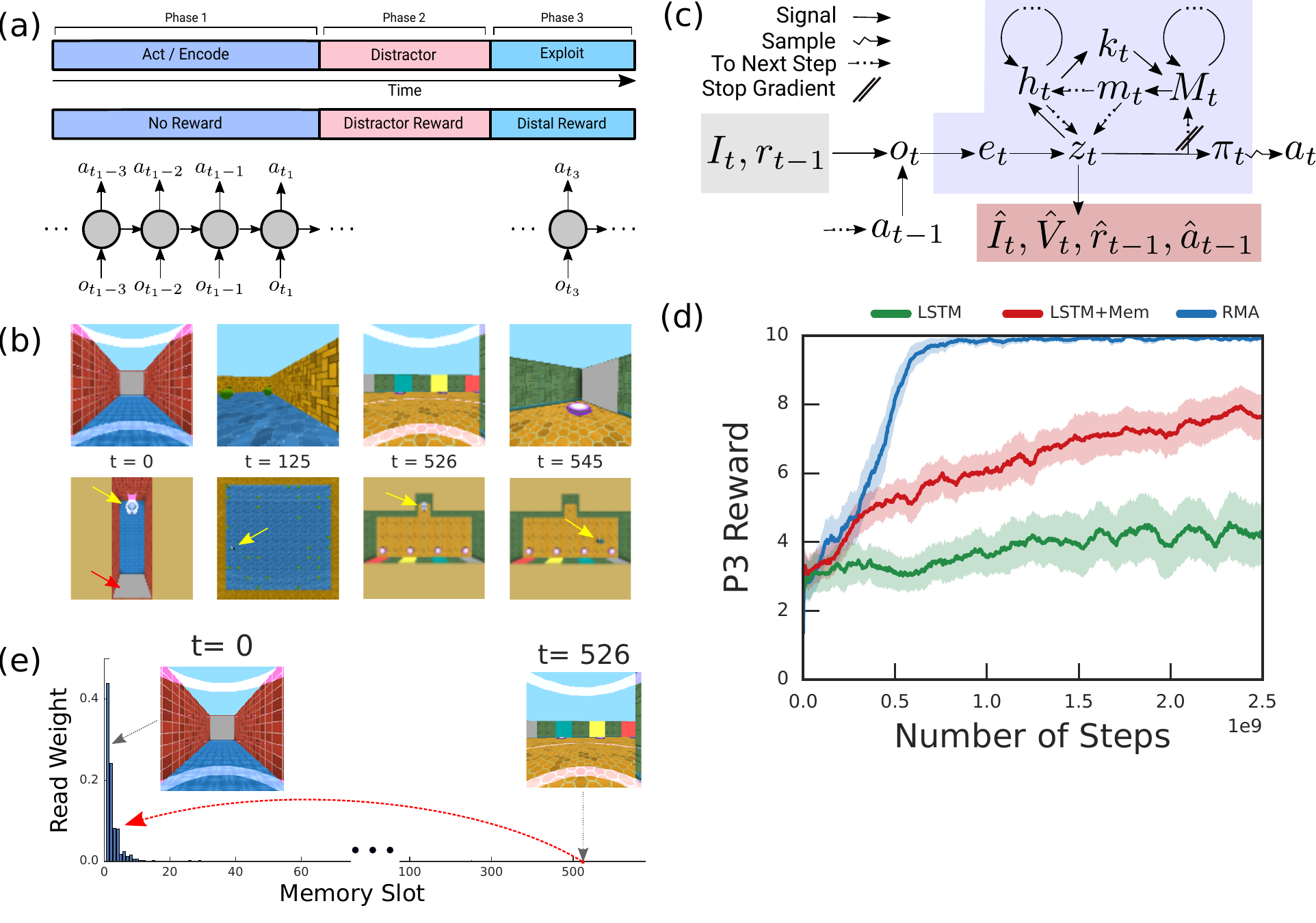}
    \caption{\textbf{Task Setting and Reconstructive Memory Agent.} \emph{a.} The three phase task structure. In phase 1 (P1), there is no reward, but the agent must seek information or trigger an event. In phase 2 (P2), the agent performs a distractor task that delivers reward. In phase 3 (P3), the agent can acquire a distal reward, depending on its behavior in P1. At each time step, the RL agent takes in observations $o_t$ and produces actions $a_t$, and passes memory state to the next time step. \emph{b.} The Passive Visual Match task: the agent passively observes a colored square on the wall in P1 (gray here), consumes apples in P2, and must select from a lineup the previously observed square from P1. The agent and colored square are indicated by the yellow and red arrow, respectively. \emph{c.} The Reconstructive Memory Agent (RMA) takes in observations, $o_t$, encodes them, $e_t$, compresses them into a state variable $z_t$, and decodes from $z_t$ the observations and value prediction $\hat{V}_t$. The state variable is also passed to an RNN controller $h_t$ that can retrieve (or read) memories $m_t$ from the external memory $M_t$ using content-based addressing with search keys $k_t$. $z_t$ is inserted into the external memory at the next time step, and the policy $\pi_t$ stochastically produces an action $a_t$ as a function of $(z_t, m_t, h_t)$ (only $z_t$ shown). \emph{d.} The RMA solves the Passive Visual Match, achieving better performance than a comparable agent without the reconstruction objective (and decoders), LSTM+Mem, and better than an agent without an external memory, LSTM. Here and henceforth, all learning curves show standard errors about the mean, computed over 5 independent runs. \emph{e.} The RMA uses its attentional read weight on time step 526 in P3 to retrieve the memories stored on the first few time steps in the episode in P1, when it was facing the colored square, to select the corresponding square and acquire the distal reward, worth 10 points.}
    \label{fig:fig1}
\end{figure}

To study decision-making in the face of long delay intervals and unrelated intervening activity, we formalize abstract task structures of two basic types. Each type is composed of three ``phases'' (Figure~1a). In the first task type (information acquisition tasks), in phase 1 (P1) the agent must, without any immediate reward, explore an environment to acquire information; in phase 2 (P2) the agent engages in an unrelated distractor task over a long time period with numerous incidental rewards; in phase 3 (P3) the agent must exploit the information acquired in P1 to succeed and acquire a \emph{distal} reward. In the second task type (causation tasks), the agent must act to trigger some event in P1 that has only long-term causal consequences. P2 is similarly a distractor task, but in P3 the agent must now exploit the changes in environment provoked by its activity in P1 to achieve success. Because a critical component of the solution we will subsequently propose involves memory encoding and retrieval, we nominally consider P1 to consist of ``action'' followed by memory encoding, P2 as the ``distractor'', and P3 as ``exploitation'' (Figure~1a). While we will sometimes report the performance in P2, e.g.~to make sure that all agents show comparable performance on the distractor task, we will focus primarily on the performance obtained by the agent in P3 as the quantity of interest. The challenge submitted to the agent is to produce behavior in P1 that assists performance in P3, thereby achieving long-term temporal credit assignment. While this task structure is contrived, it enables us to systematically control delay durations and variance in the distractor reward.

Under the assumptions of this task structure, we can understand why a distractor phase can be particularly damaging to long-term temporal credit assignment by defining a measure of the signal-to-noise ratio (SNR) in the policy gradient estimate that induces behavioral adaptation in P1. Here, we measure the SNR as the squared length of the expected gradient, $\|\E_\pi[\Delta \theta]\|^2$, divided by the variance of the gradient estimate, $\text{Var}_\pi[\Delta \theta]$ (which is the trace of $\text{Cov}_\pi(\Delta \theta, \Delta \theta)$). In Supplement Section 6, we show that with $\gamma=1$ the SNR is approximately
\begin{align}
\mathrm{SNR} & \approx \frac{\displaystyle \|\E_\pi [\Delta \theta]\|^2}{\displaystyle \text{Var}_\pi \big [ \sum_{t \in P2} r_{t} \big ] \times C(\theta) + \text{Var}_\pi[\Delta \theta | \text{no P2}]},  
\end{align}
where $C(\theta)$ is a reward-independent term, and $ \text{Var}_\pi[\Delta \theta | \text{no P2}]$ is the (trace of the) policy gradient variance in an equivalent problem without a distractor interval. $\text{Var}_\pi \big [ \sum_{t \in P2} r_{t} \big ]$ is the reward variance in P2. When P2 reward variance is large, the policy gradient SNR is inversely proportional to it. Reduced SNR is known to adversely affect the convergence of stochastic gradient optimization\cite{roberts2009signal}. The standard solution is to average over larger data batches, which, with independent samples, linearly increases SNR. However, this is necessarily at the expense of data efficiency and becomes more difficult with longer delays and more interceding variance.

Before we examine a complete task of this structure, consider a simpler, related task, which we call Passive Visual Match (Figure~1b), that involves a long time delay and memory dependence without long-term temporal credit assignment. This task is \emph{passive} in that the information that must be remembered by the agent is observed passively without any action required on its part; tasks of this form have been recently studied in memory-based RL\cite{wayne2018unsupervised, ritter2018been}. In Passive Visual Match, the agent begins each episode in a corridor facing a wall with a painted square whose color is set at random. While this corresponds to the period P1 in the task structure, the agent does not need to achieve any goal here. After five seconds, the agent is transported to another room in which it engages in the distractor task of collecting apples for a 30 second period in P2. Finally, in P3 the agent is transported to a third room in which four colored squares are posted on the back wall, one of which matches the observation in P1. If the agent moves to the groundpad in front of the matching colored square, it receives a distal reward, which is in fact much smaller than the total distactor phase reward. To solve this task, it is unnecessary for the agent to take into account reward from the distant future to make decisions as the actions in P3 precede reward by a short interval. However, the agent must be able to store and access memories of its past: here, it must memorize the P1 color cue, maintain that information over the P2 interval, and retrieve it to choose a pad. 

\section*{The Reconstructive Memory Agent}

We solve this task with a vision and memory-based agent, which we name the \emph{Reconstructive Memory Agent} (RMA) (Figure~1c), which is based on a previously published agent model\cite{wayne2018unsupervised} but simplified for the present study. Critically, this agent model combines a reconstruction process to compress useful sensory information with memory storage that can be queried by content-based addressing\cite{bahdanau2014neural, graves2014neural, graves2016hybrid} to inform the agent's decisions. The RMA itself does not have specialized functionality to subserve long-term temporal credit assignment but provides a basis for the operation of the Temporal Value Transport algorithm, which does.

In this model, an image frame $I_t$, the previous reward $r_{t-1}$, and the previous action $a_{t-1}$ constitute the observation $o_t$ at time step $t$. These inputs are processed by encoder networks and merged into an embedding vector $e_t$, which is to be combined with the output of a recurrent neural network (RNN) based on the Differentiable Neural Computer\cite{graves2016hybrid}. This RNN consists of a recurrent LSTM ``controller'' network $h$ and a memory matrix $M$ of dimension $N \times W$. The output of this RNN and memory system from the previous time step $t-1$ consists of the LSTM output $h_{t-1}$ and $k$ ($=3$ here) vectors of length $W$ read from memory $m_{t-1} \equiv (m_{t-1}^{(1)}, m_{t-1}^2, \dots, m_{t-1}^{(k)})$, which we refer to as memory read vectors. Together, these outputs are combined with the embedding vector by a feedforward network into a ``state representation'' $z_t = f(e_t, h_{t-1}, m_{t-1})$. Importantly, the state representation $z_t$ has the same dimension $W$ as a memory read vector. Indeed, once produced it will be inserted into the memory at the next time step into the $t$-th row: $M_{t+1}[t, \cdot] \leftarrow z_t$. 

Before this occurs, however, the RNN carries out one cycle of reading from memory and computation. The state representation $z_t$ is provided as input to the RNN, alongside the previous time step's memory read vectors $m_{t-1}$ to produce the next $h_t$. Then reading memory to produce the current time step's memory read vectors occurs: $k$ read keys $k_t^{(1)}, k_t^{(2)}, \dots, k_t^{(k)}$ of dimension $W$ are produced as a function of $h_t$, and each key is matched against every row $n$ using a similarity measure $S(k_t^{(i)}, M_{t-1}[n,\cdot])$. 
The similarities are scaled by a positive read strength parameter $\beta_t^{(i)}$ (also computed as a function of $h_t$), to which a softmax over the weighted similarities is applied.
This creates an attentional read weight vector $w_t^{(i)}$ with dimension $N$, which is used to construct the $i$-th memory read vector $m_t^{(i)} = \sum_{n=1}^N w_t^{(i)}[n] M_t[n, \cdot]$.

The state representation $z_t$ is also sent to decoder networks whose objective functions require them to produce reconstructions $\hat{I}_t, \hat{r}_{t-1}, \hat{a}_{t-1}$ of the observations (the carets denote approximate quantities produced by networks) while also predicting the value function $\hat{V}(z_t)$. This process ensures that $z_t$ contains useful sensory information in a compressed format. Finally, the state representation $z_t$ and RNN outputs $(h_t,m_t)$ are provided as input to the policy network to construct the policy distribution $\pi(a_t | z_t, h_t, m_t)$, which is a multinomial distribution over the discrete action space here. At each time step, an action $a_t$ is sampled and applied to the environment.

When trained on Passive Visual Match, all the agents we tested did succeed at the apple collection distractor task (Supplementary Figure~1), although only the RMA learned to solve the distal reward task by appropriately selecting the same colored square in P3 as was seen in P1 (Figure~1d). A comparison agent without an external memory (the LSTM agent) was able to achieve only slightly better than chance performance in P3, and a comparison agent with an external memory but no reconstruction objective decoding observation data from $z_t$ (the LSTM+Mem agent) also performed worse. The reconstruction process in the RMA helps to build and stabilize perceptual features in $z_t$ that can later be found by memory retrieval\cite{wayne2018unsupervised}. The solution of the RMA was robust. In Supplementary Figure 2, we demonstrate equivalent results for 0, 15, 30, 45, and 60 second distractor intervals: the number of episodes required to learn remained roughly independent of the delay (Supplementary Figure 3). Additionally, for more complicated visual stimuli consisting of CIFAR images\cite{krizhevsky2014cifar}, the RMA was also able to make correct matching choices (Supplementary Figure 4).

Despite the delay between P1 and P3, Passive Visual Match does not require long-term temporal credit assignment. The cue in P1 is automatically observed; an agent only needs to encode and retrieve a memory to move to the correct pad in P3 -- a process that is relatively brief. Consequently, an agent with a small discount factor $\gamma = 0.96$ ($\tau = 25$ steps at $15$ frames per second, giving a 1.67 second half-life) was able to solve the task. However, the ability to encode and attend to specific past events was critical to the RMA's success. In Figure 1e, we see the attentional weighting vector $w_t$ produced by one of the RMA read keys in an episode at time step 526, which corresponds to the beginning of P3. The weighting was sparsely focused on memories written in the first few episode time steps, during the instants when the agent was encoding the colored square. The learned memory retrieval identified relevant historical time points and bridged the 30 second distractor interval. Recall of memories in the RMA is driven by the demand of predicting the value function $\hat{V}(z_t)$ and producing the policy distribution $\pi(a_t | z_t, h_t, m_t)$. As we have seen, these objectives allowed the agent to automatically detect past time points that were relevant to its current decision. 

\begin{figure}
    \centering
    \includegraphics[width=0.99\textwidth]{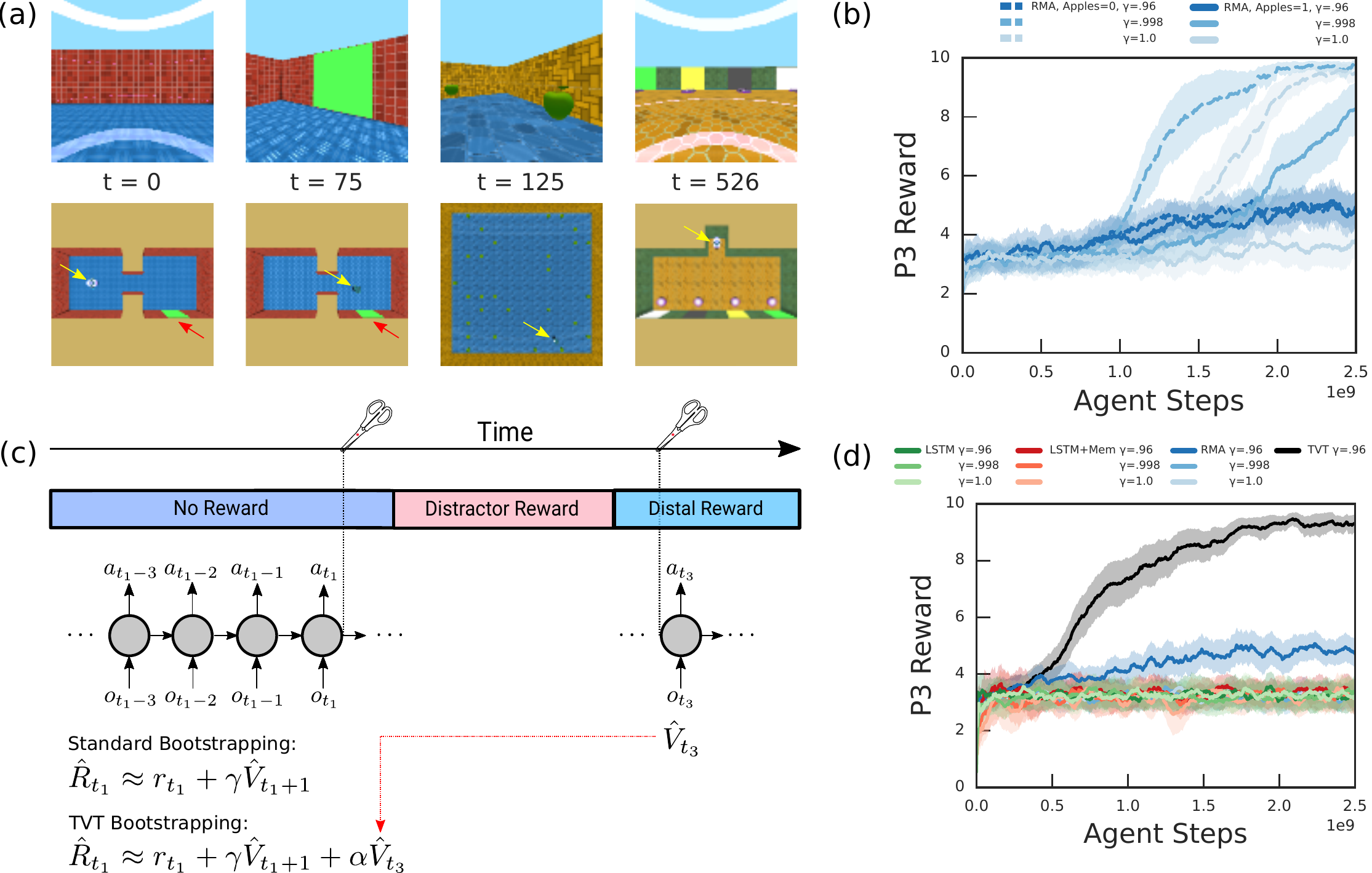}
    \caption{\textbf{Temporal Value Transport and Type 1 Information Acquisition Tasks.} \emph{a.} First-person (upper row) and top-down view (lower row) in Active Visual Match task while the agent is engaged in the task. In contrast to Passive Visual Match, the agent must explore to find the colored square, randomly located in a two-room environment. The agent and colored square are indicated by the yellow and red arrow, respectively. \emph{b.} Without rewards in P2, RMA models with large discount factors (near 1) were able to solve the task; the RMA with $\gamma=0.998$ exhibited retarded but definite learning with modest P2 reward (1 point per apple). \emph{c.} Cartoon of the Temporal Value Transport mechanism: the distractor interval is spliced out, and the value prediction $\hat{V}_{t_3}$ from a time point $t_3$ in P3 is directly added to the reward at time $t_1$ in P1. \emph{d.} The TVT agent alone was able to solve Active Visual Match with large rewards during the P2 distractor, and faster than agents exposed to no distractor reward. The RMA with discount factor $\gamma=0.96$ was able to solve a greater than chance fraction because it could randomly encounter the colored square in P1 and retrieve its memory in P3.}
    \label{fig:fig2}
\end{figure}

We now turn to a type 1 information acquisition task, Active Visual Match, that truly demands long-term temporal credit assignment. Here, in P1 the agent must actively seek out a colored square, randomly located in a two-room maze, so that it can accurately decide on the match in P3 (Figure 2a). If an agent finds the visual cue by chance in P1, then it can use this information in P3, but this will only be successful at random. As in Passive Visual Match, the agent engages in a 30 second distractor task of apple collection during P2. When the rewards of P2 apples were set to 0, RMAs with discount factors sufficiently close to 1 were able to solve the task (Figure~2b, dashed lines). With a randomized number of apples worth one point each, the RMAs with $\gamma=0.998$ ultimately began to learn the task (Figure~2b, solid line, medium blue) but were slower in comparison to the no P2 reward case. For a fixed mean reward per episode in P2 but increasing variance, RMA agent performance degraded entirely (Supplementary Figure 5). Finally, for the principal setting of the level, where each P2 apple is worth five points, and the P2 reward variance is $630$, all comparison models (the LSTM agent, LSTM+Mem agent, and RMA) failed to learn P1 behavior optimized for P3 (Figure~2d). For $\gamma = 0.96$, RMAs reached a score of about 4.5, which implies slightly better than random performance in P3: this was because RMAs solved the task in cases where they accidentally sighted the cue in P1.

\section*{Temporal Value Transport}

Temporal Value Transport (TVT) is a learning algorithm that augments the capabilities of memory-based agents to solve long-term temporal credit assignment problems. The insight behind TVT is that we can combine attentional memory access with reinforcement learning to fight variance by automatically discovering how to ignore it, effectively transforming a problem into one with no delay at all. A standard technique in RL is to estimate the return for the policy gradient calculation by \emph{bootstrapping}\cite{sutton1998reinforcement}: using the learned value function, which is deterministic and hence low variance but biased, to reduce the variance in the return calculation. We denote this bootstrapped return as $\tilde{R}_t := r_t + \gamma \hat{V}_{t+1}$. The agent with TVT (and the other agent models considered here) likewise bootstraps from the next time step and uses a discount factor to reduce variance further. However, it additionally bootstraps from the distant future.

\begin{figure}
    \centering
    \includegraphics[width=0.99\textwidth]{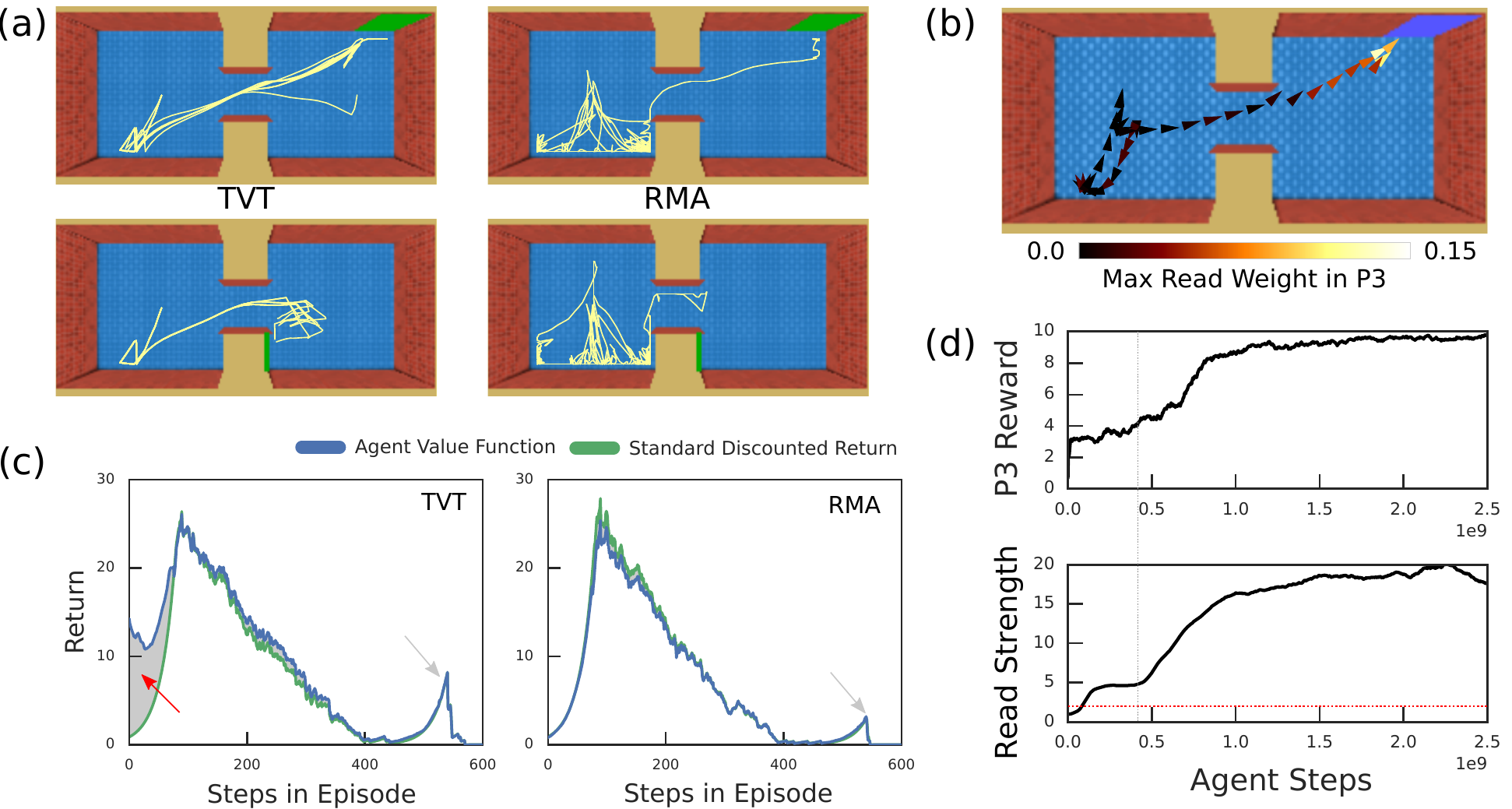}
    \caption{\textbf{Analysis of Agent in Active Visual Match.} \emph{a.} In P1, TVT trained on Active Visual Match, actively sought out and oriented to the colored squared. RMA meandered randomly. \emph{b.} Its attentional read weights focused maximally on the memories from time points when it was facing the colored square. \emph{c.} With statistics gathered over 20 episodes, TVT's average value function prediction in P1 (blue) was larger than the actual discounted reward trace (green) -- due to the transported reward. Difference shown in gray. The RMA value function in contrast matched the discounted return very closely. \emph{d.} The P3 rewards for TVT rose during learning (upper panel) after the maximum read strength per episode first crossed threshold on average (lower panel, red line).}
    \label{fig:fig3}
\end{figure}

In Figure~2c, we highlight the basic principle behind TVT. We previously saw in the Passive Visual Match task that the RMA reading mechanism learned to retrieve a memory from P1 in order to produce the value function prediction and policy in P3. This was a purely automatic process determined by the needs of the agent in P3.
We exploit this phenomenon to form a link between the time point $t_3$ (occurring, for example, in P3) at which the retrieval occurs and the time $t_1$ at which the retrieved memory was encoded. This initiates a \emph{splice event} in which the bootstrapped return calculation for $t_1$ is revaluated to $\tilde{R}_{t_1} := r_{t_1} + \gamma \hat{V}_{t_1 + 1} + \alpha \hat{V}_{t_3}$, where $\alpha$ is a form of discount factor that diminishes the impact of future value over multiple stages of TVT. From the perspective of learning at time $t_1$, the credit assignment is conventional: the agent tries to estimate the value function prediction based on this revaluated bootstrapped return, and it calculates the policy gradient based on it as well. The bootstrapped return can trivially be regrouped as $\tilde{R}_{t_1} : = (r_{t_1} + \alpha \hat{V}_{t_3}) + \gamma \hat{V}_{t_1 + 1}$, which facilitates the interpretation of the transported value as a fictitious reward introduced to time step $t_1$. 

\begin{algorithm}[ht]
\caption{Temporal Value Transport for One Read}
\begin{algorithmic}
    \STATE \textbf{input:} $\{r_t\}_{t \in [1,T]}$, $\{\hat{V}_t\}_{t \in [1,T]}$, 
    read strengths $\{\beta_t\}_{t \in [1,T]}$, 
    read weights $\{w_t\}_{t \in [1,T]}$
    \STATE splices : = [] 
    \FOR{\textbf{each} crossing of read strength $\beta_t$ above $\beta_\text{threshold}$}
        \STATE $t_\text{max} : = \arg \max_t \{\beta_t | t \in \text{crossing window} \}$
        \STATE Append $t_\text{max}$ to splices 
    \ENDFOR
        \FOR{$t$ in 1 to T}       
            \FOR{$t'$ in splices}
                \IF{$t < t' - 1/(1-\gamma)$} 
                    \STATE $r_t := r_t + \alpha w_{t'}[t] \hat{V}_{t'+1}$\\ 
                    \COMMENT{The read based on $w_{t'}$ influences value prediction at next step, hence $\hat{V}_{t'+1}$.} 
                \ENDIF
            \ENDFOR
        \ENDFOR
    \RETURN $\{r_t\}_{t \in [1,T]}$
\end{algorithmic}
\label{alg:train}
\end{algorithm}

This characterization is broadly how TVT works. However, in detail there are multiple practical issues to understand further. First, the TVT mechanism only triggers a splice event when a memory retrieval is sufficiently strong: in particular, this occurs whenever a read strength $\beta_{t}^{(i)}$ is above a threshold value, $\beta_\text{threshold}$. Second, each of the $k$ memory reading processes operates in parallel, and each can independently trigger a splice event. Third, instead of linking to a single past event, the value at the time of reading $t'$ is transported back to all time points $t$ with a strength proportional to the attentional weighting $w_{t'}[t]$. Fourth, value is not transported to events that occurred very recently, where recently is any time within one half-life $\tau = 1 / (1-\gamma)$ of the reading time $t'$. Pseudocode for the TVT algorithm is shown in Algorithm \ref{alg:train}, and further implementation details are discussed in Supplement Section~5.

When applied to the Active Visual Match task with large distractor reward, an RMA model equipped with TVT (henceforth just TVT) learned the behavior in P1 that produced distal reward in P3; it also learned the task faster than did any RMA with no distractor reward (Figure 2b\&d). The difference in learned behavior was dramatic: TVT reliably sought out and oriented toward the colored square in P1, while the standard RMA behaved randomly (Figure 3a). Figure 3b overlays on the agent's trajectory (arrowheads) a coloring based on the read weight produced at the time $t_3$ of a TVT splice event in P3: TVT learned to read effectively from memories in P1 associated with the time points for which it was viewing the colored square. During the learning process, we see that the maximum read strength recorded per episode (Figure 3d, lower panel) began to reach threshold (lower panel, red line) early and prior to producing P3 reward reliably (Figure 3d, upper panel), which then instigated the learned behavior in P1. After training, TVT's value function prediction $\hat{V}_t$ directly reflected the fictitious rewards. Averaged over 20 trials, the value function in P1 (Figure 3c, left panel, blue curve) was higher than the actual discounted return, $\sum_{t' \geq t} \gamma^{t'-t} r_{t'}$, (Figure 3c, left panel, green curve). The RMA agent with discounting did not show a similar difference between the discounted return and the value function (Figure 3c, right panel). In both Figure 3c panels, we see bumps in P3 in the return traces due to the distal reward: TVT achieved higher reward in general, with the RMA return reflecting only chance performance. Further, we examined whether TVT could solve problems with even longer distractor intervals, in this case with a P2 interval of 60 seconds. TVT also learned here (Supplementary Figure 6).

\begin{figure}
    \centering
    \includegraphics[width=0.99\textwidth]{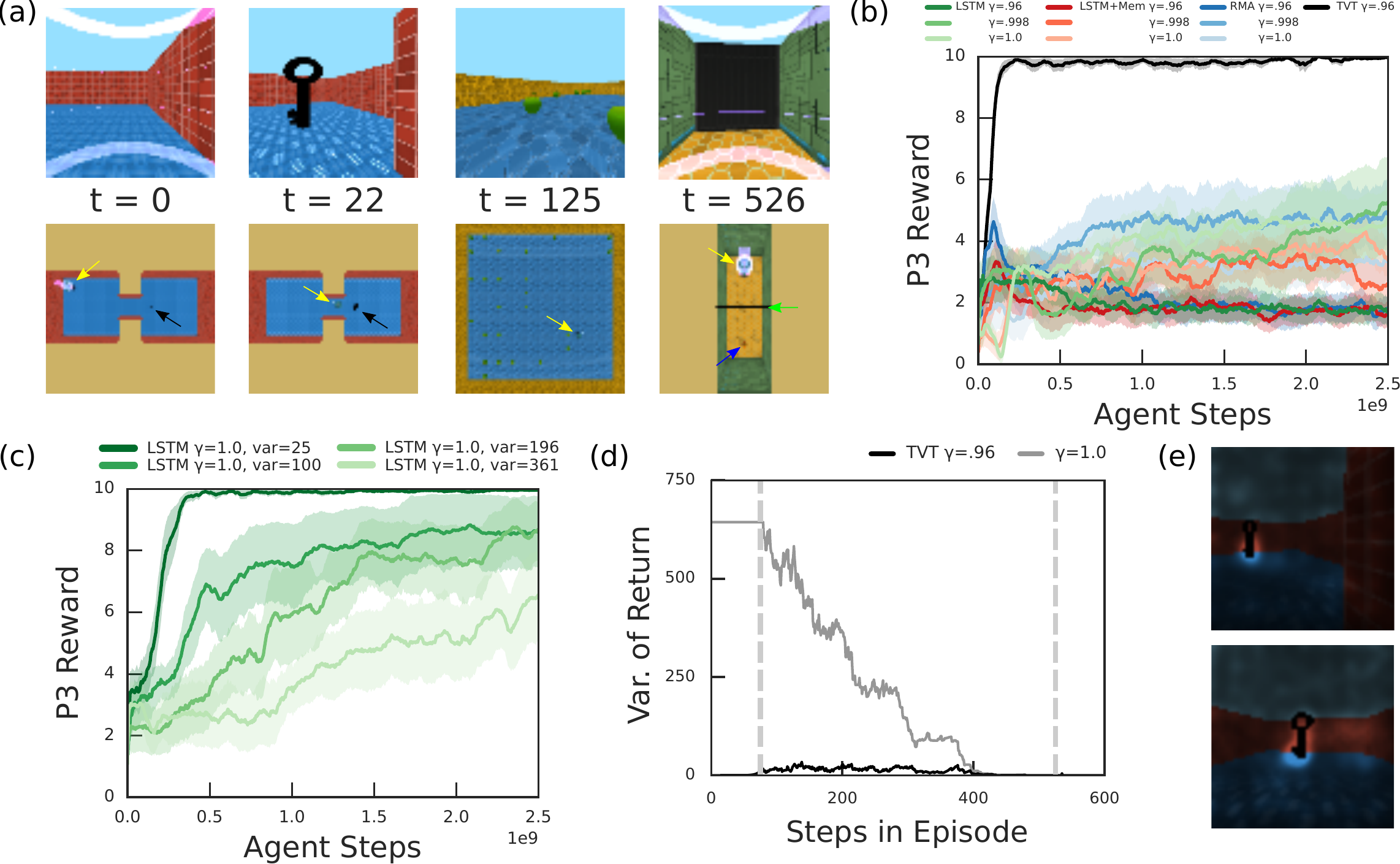}
    \caption{\textbf{Type 2 Causation Tasks.} \emph{a.} First person (upper row) and top-down view (lower row) in Key-to-Door task. The agent (indicated by yellow arrow) must pick up a key in P1 (black arrow), collect apples in P2, and, if it possesses the key, it can open the door (green arrow) in P3 to acquire the distal reward (blue arrow). \emph{b.} Learning curves for P3 reward (TVT in black). Although this task requires no memory for the policy in P3, computing the value prediction still triggers TVT splice events, which promote key retrieval in P1. \emph{c.} Increasing the standard deviation of reward available in P2 disrupted the performance of LSTM agents at acquiring the distal reward. \emph{d.} On 20 trials produced by a TVT agent, we compared the variance of the TVT bootstrapped return against the undiscounted return. The TVT return's variance was orders of magnitude lower. Vertical lines mark phase boundaries. \emph{e.} Saliency analysis of the pixels in the input image in P1 that the value function gradient is sensitive to. The key pops out in P1.}
    \label{fig:fig4}
\end{figure}

Temporal Value Transport can also solve type 2 causation tasks, where the agent does not need to acquire information in P1 for P3 but instead must cause an event that will affect the state of the environment in P3. Here, we study the Key-to-Door (KtD) task in which an agent must learn to pick up a key in P1 so that it can unlock a door in P3 to obtain reward (Figure~4a). Although no information from P1 must be recalled in P3 to inform the policy's actions (the optimal decision is to move toward the door in P3 regardless of the events in P1), TVT still learned to acquire the key in P1 because it read from memory to predict the value function when positioned in front of the door in P3 (Figure~4b, black), while all other agents failed to pick up the key reliably in P1 (Figure~4b blue, pink, green). In this case, the P2 reward variance was comparatively low -- with the only variance due to a randomized number of apples but with each apple consistently giving $r_\text{apple} = 5$. In higher SNR conditions (low P2 reward variance), even LSTM agents with $\gamma=1$ were able to solve the task, indicating that a large memory itself is not the primary factor in task success (Figure 4c). TVT specifically assisted in credit assignment. However, the LSTM agents could learn only for small values of P2 reward variance, and performance degraded predictably as a function of increasing reward variance in P2 (Figure 4c, dark to light green curves). For the same setting as Figure~4b, we calculated the variance of either the TVT bootstrapped return $\tilde{R}_t$ for each time point, over 20 episodes, and compared on the same episodes to the variance of the undiscounted return, $\sum_{t' \geq t} r_{t'}$ (Figure 4d). 
Because it exploits discounting, the variance of the bootstrapped return of TVT was nearly two orders of magnitude smaller in P1. We next asked if the agent attributed the fictitious reward transported to P1 in an intelligent way to the key pickup. In P1, using a saliency analysis similar to\cite{simonyan2013deep}, we calculated the gradient of the value function prediction with respect to the input image $\nabla_{I_t} \hat{V}_t(z_t)$ and shaded the original input image in proportion to the magnitude of this quantity (Supplement Section 8.2). In Figure 4e, we see that this produced a segmentation of the key, indicating that the P1 value prediction was most sensitive to the observation of the key. As a control experiment, in Supplementary Figure 7, we tested if there needed to be any surface-level similarity between visual features in P3 and the encoded memory in P1 for memory retrieval to function. With a blue instead of a black key, TVT also solved the task as easily, indicating that the memory searches could flexibly find information with a somewhat arbitrary relationship to current context.

One can understand how TVT learned to solve this task as a progression. Initially, on a small fraction of the episodes, the agent picked up the key at random. From this point, the agent learned, on encountering the door, to retrieve memories from P1 that identified if the agent picked up the key in order to predict the return in P3 accurately (this is what RMA did as well). Whenever the memories from P1 were retrieved, splice events were triggered that transported value back to the behavioral sequences in P1 that led to key pickup. 

\begin{figure}
    \centering
    \includegraphics[width=0.99\textwidth]{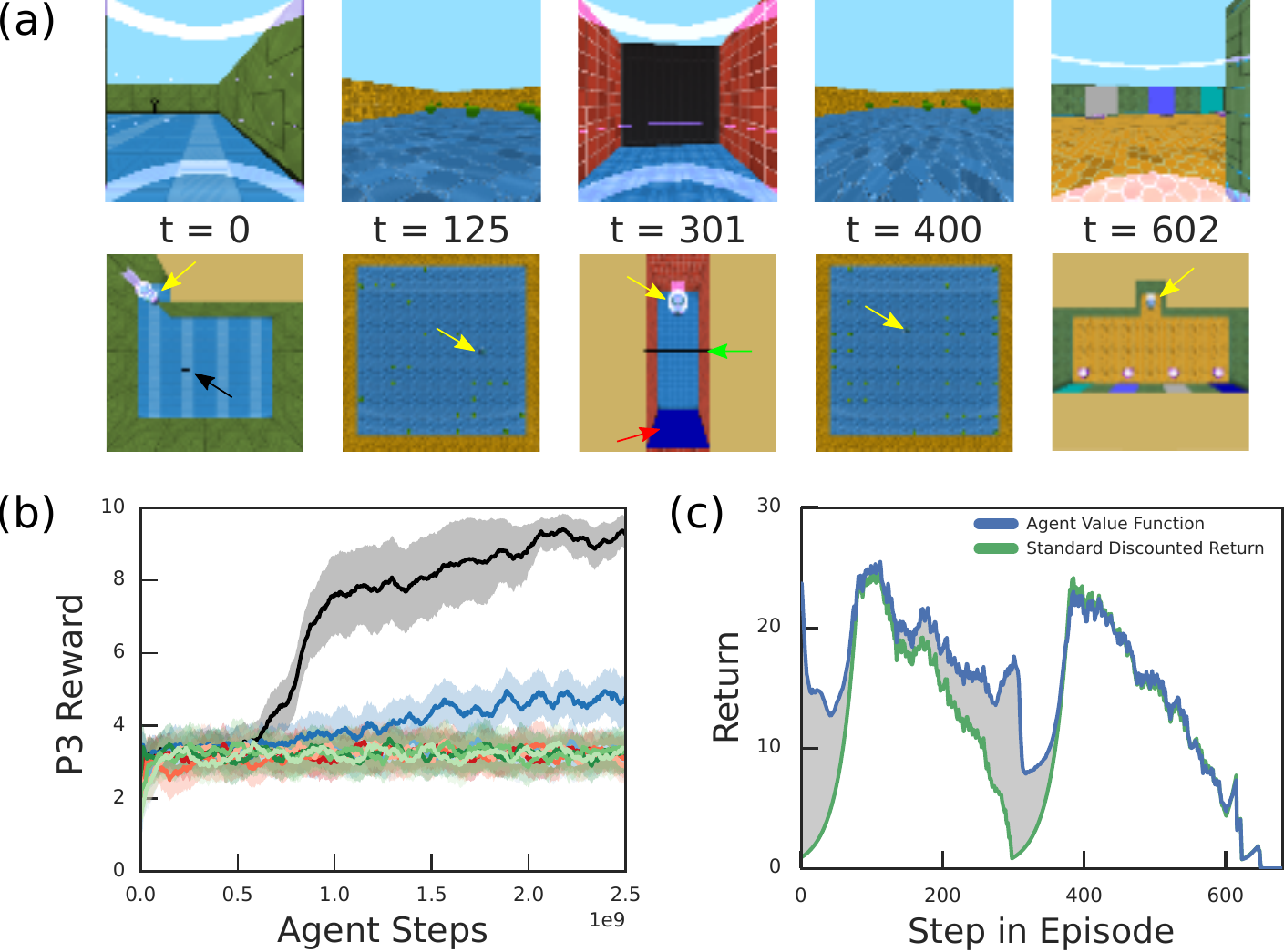}
    \caption{\textbf{Transport across Multiple Phases.} \emph{a.} Key-to-Door-to-Match (KtDtM) task. The agent (yellow arrow) must pick up a key (black arrow) in P1, to open a door (green arrow) and encode a colored square (red arrow) in P3, to select the matching colored square in P5. P2 and P4 are distractor apple collecting tasks. \emph{b.} TVT (black) solved this task, whereas RMA (blue) solved the P5 component of the task when it by chance retrieved the P1 key and opened the door in P3. \emph{c.} The value function prediction (blue) in TVT developed two humps where it was above the discounted return trace (green), one in P1, one in P3, encoding the value of achieving the ``sub-goals'' in P1 and P3.}
    \label{fig:fig5}
\end{figure}

The introduction of transported value can come at a cost. When a task has no actual need for long-term temporal credit assignment, spurious triggering of splice events can send value back to earlier time points and bias the agent's activity. To study this issue, we examined performance of TVT on a set of independently developed RL tasks that were designed in a context where standard discounted RL was expected to perform well. We compared the performance of the LSTM agent, the LSTM+Mem agent, RMA, and TVT. TVT generally performed on par with RMA on many tasks but slightly worse on one (Supplementary Figures~8-9) and outperformed all of the other agent models, including LSTM+Mem. We also considered whether TVT would function when P3 reward was strictly negative, but a behavior in P1 could be developed to avert a larger disaster. In the Two Negative Keys task, the agent is presented with a blue key and red key in a room in P1. If the agent picks up the red key, it will be able to retrieve a distal reward behind a door in P3 worth $-1$; if it picks up the blue key, it will be able to retrieve a distal reward worth $-10$, and if it does not pick up a key at all, it is penalized $-20$ in P3. TVT was also able to solve this task (Supplementary Figure~10). 

Having established that TVT was able to solve relatively simple problems, we now demonstrate TVT's capability in two more complex scenarios. The first of these is an amalgam of the KtD and the Active Visual Match task, which demonstrates temporal value transport across multiple phases -- the Key-to-Door-to-Match task (KtDtM); here, an agent must exhibit two non-contiguous behaviors to acquire the distal reward. 

In this task, instead of a three phase structure, we have five phases: P1-P5 (Figure 5a). P2 and P4 are both long distractor phases involving apple collection distractor rewards. In P1 and P3, there are no rewards. In P1, the agent must fetch a key, which it will use in P3 to open a door to see a colored square. In P5, the agent must choose the groundpad in front of the colored square matching the one that was behind the door in P3. If the agent does not pick up the key in P1, it is locked out of the room in P3 and cannot make the correct choice in P5. TVT solved this task reliably (Figure 5b), whereas all other agents solved this problem only at chance level in P5, and did not pursue the key in P1. As might be expected, the TVT value function prediction rose in both P1, P3, and P5 (Figure 5c) with two humps where the P1 and P3 value functions were above the discounted return traces. Because the discount factor $\alpha$ for TVT transport was relatively large (0.9), the two humps in the value prediction were of comparable magnitude.

\begin{figure}
    \centering
    \includegraphics[width=0.99\textwidth]{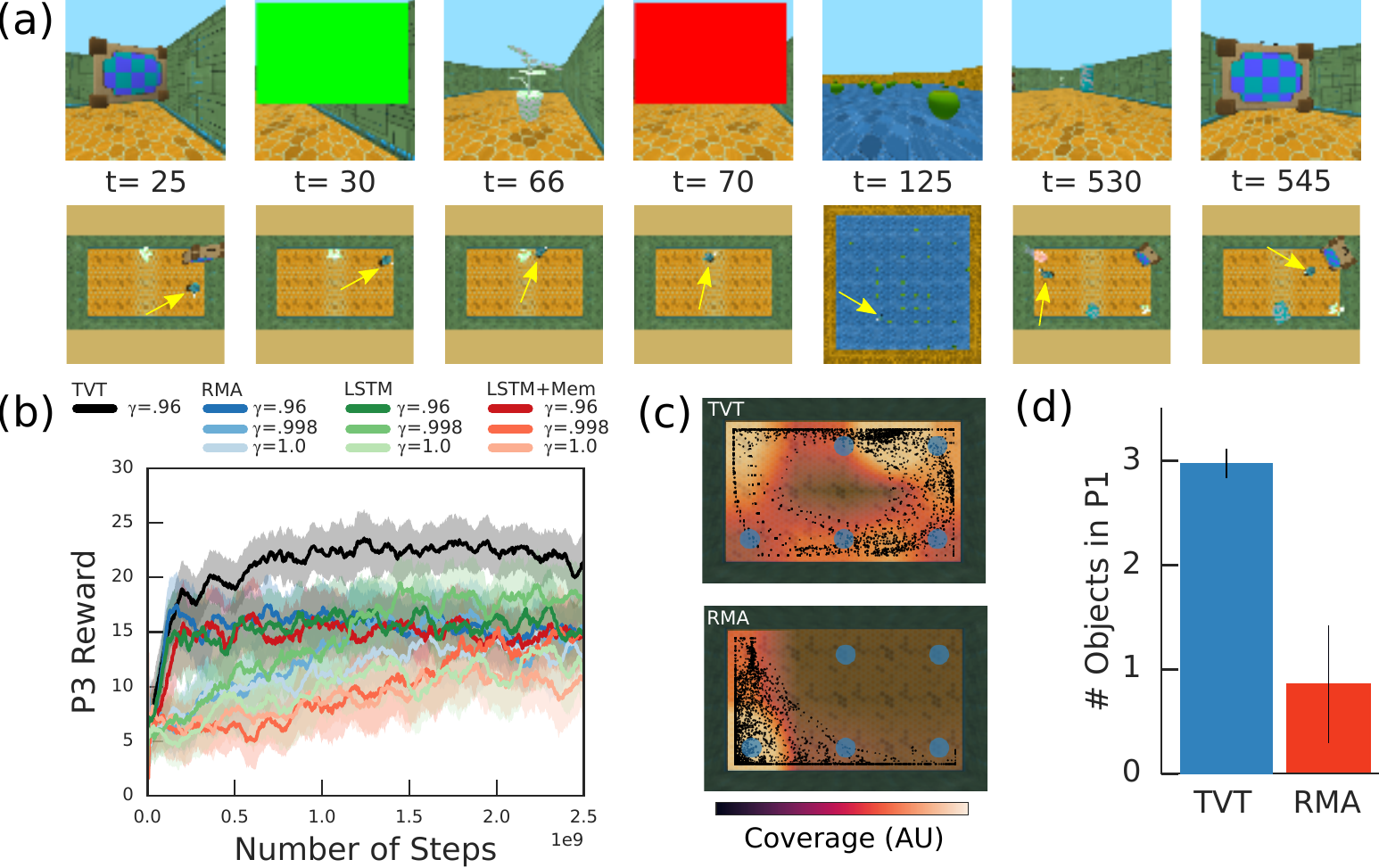}
    \caption{\textbf{More Complex Information Acquisition.} \emph{a.} In Latent Information Acquisition, the agent (yellow arrow) must touch three procedurally generated objects to identify from a subsequent color flash if each is either green or red. In P3, green objects yield positive reward and red objects negative. \emph{b.} TVT performed well on this task (black curve). \emph{c.} In 20 trials, we plot the positional coverage in P1 of a TVT agent compared to RMA. TVT developed exploratory behavior in P1: it navigated among the six possible locations where the P1 objects could be placed, whereas the RMA typically moved into the corner. \emph{d.} A quantification over 20 trials of the exploratory behavior in P1: TVT usually touched all three of the objects in P1, whereas RMA touched about one.}
    \label{fig:fig6}
\end{figure}

Finally, we look at a richer information acquisition task, Latent Information Acquisition (Figure 6a). In P1, the agent begins in a room surrounded by three objects with random textures and colors drawn from a set. During P1, each object has no reward associated with it. When an object is touched by the agent, it disappears and a color swatch (green or red) appears on the screen. Green swatches indicate that the object is good, and red swatches indicate it is bad. The number of green- and red-associated objects was balanced on average. In P2, the agent again collects apples for 30 seconds. In P3, the agent must collect only the objects that were associated with a green swatch. 

The TVT agent alone was able to solve the task (Figure 6b, black curve), usually touching all three objects in P1 (Figure 6d), while the RMA only touched one object on average, and it outperformed non-TVT agents by a wide margin (Figure 6b, other colors).
The non-TVT agents all exhibited pathological behavior in P1. In P1, the objects were situated on a grid of six possible locations (with no relationship to P3 location). TVT learned an exploratory sweeping behavior whereby it efficiently covered the locations where the objects were present (Figure 6c), whereas RMA reliably moved into the same corner, thus touching by accident only one object. 

\section*{Discussion}

The mechanism of TVT should be compared to other recent proposals to address the problem of long-term temporal credit assignment. The Sparse Attentive Backtracking algorithm\cite{ke2018sparse} in a supervised learning context uses attentional mechanisms over the states of an RNN to propagate backpropagation gradients effectively. The idea of using attention to the past is shared with our work; however, there are substantial differences. Instead of propagating gradients to shape network representations, in the RMA we have used temporally local reconstruction objectives to ensure relevant information is encoded and stored in the memory. Further, backpropagating gradients to RNN states would not actually train a policy's action distribution, which is the crux of reinforcement learning. Our approach instead modifies the rewards from which the full policy gradient is derived. Like TVT, the RUDDER algorithm\cite{arjona2018rudder} has recently been proposed in the RL context to address the problem of learning from delayed rewards. RUDDER uses an LSTM to make predictions about future returns and sensitivity analysis to decompose those returns into reward packets distributed throughout the episode. TVT is explicitly designed to use a reconstructive memory system to compress high-dimensional observations in partially-observed environments and retrieve them with content-based attention. At present, we know of no other algorithm that can solve type 1 information acquisition tasks.

Temporal Value Transport is a heuristic algorithm but one that expresses coherent principles we believe will endure: past events are encoded, stored, retrieved, and revaluated. TVT fundamentally intertwines memory systems and reinforcement learning: the attention weights on memories specifically modulate the reward credited to past events. While not intended as a neurobiological model, the notion that neural memory systems and reward systems are highly co-dependent is supported by much evidence, including the existence of direct dopaminergic projections to hippocampal CA1 and the contribution of D1/D5 dopamine receptors in acquiring task performance in awake-behaving animals\cite{li2003dopamine, lemon2006dopamine}. 

Throughout this work, we have seen that standard reinforcement learning algorithms are compromised when solving even simple tasks requiring long-term behavior. We view discounted utility theory, upon which almost all reinforcement learning is predicated, as the ultimate source of the problem, and our work provides evidence that other paradigms are not only possible but can work better. In economics, paradoxical violation of discounted utility theory has occasioned bountiful scholarship and diverse, incompatible, and incomplete theories\cite{frederick2002time}. We hope that a cognitive mechanisms approach to understanding ``inter-temporal choice'' -- in which preferences and long-term economic behavior are decoupled from a rigid discounting model -- will inspire new ways forward. The principle of splicing together remote events based on episodic memory access may offer a promising vantage from which to begin future study of these issues. 

The complete explanation of the remarkable ability of human beings to problem solve and express coherent behaviors over long spans of time remains a profound mystery about which our work only provides a smattering of insight. TVT learns slowly, whereas humans are at times able to discover causal connections over long intervals quickly (albeit sometimes inaccurately). Human cognitive abilities are often conjectured to be fundamentally more model-based than the mechanisms in most current reinforcement learning agents (TVT included)\cite{hassabis2017neuroscience} and can provide consciously available causal explanations\cite{pearl2018book} for events. When the book is finally written on the subject, it will likely be understood that long-term temporal credit assignment recruits nearly the entirety of the human cognitive apparatus, including systems designed for prospective planning, abstract, symbolic, and logical reasoning, commitment to goals over indefinite intervals, and language. Some of this human ability may well require explanation on a different level of inquiry altogether: among different societies, attitudes and norms regarding savings rates and investment vary enormously\cite{guyer1997endowments}. There is in truth no upper limit to the time horizons we can conceptualize and plan for.

\section*{Correspondence}
Correspondence should be addressed to Greg Wayne, Chia-Chun Hung, or Timothy Lillicrap (email: $\{$gregwayne, aldenhung, countzero$\}$@google.com).

\section*{References}

\bibliographystyle{naturemag}
\bibliography{scibib}

\newpage

\begin{center}
\textbf{\Large{\centering{Supplement for \\ Optimizing Agent Behavior over Long Time Scales by Transporting Value
}}}
\end{center}

\renewcommand{\figurename}{Supplementary Figure}
\renewcommand{\tablename}{Supplementary Table}
\setcounter{figure}{0}    
\setcounter{table}{0} 
\setcounter{tocdepth}{4}
\setcounter{secnumdepth}{4}
\tableofcontents

\newpage

\section{Agent Model}

At a high level, the Reconstructive Memory Agent (RMA) consists of four modules: an \textbf{encoder} for processing observations at each time step; a \textbf{memory augmented recurrent neural network}, which contains a deep LSTM ``controller'' network and an external memory that stores a history of the past; its output combines with the encoded observation to produce a state variable representing information about the environment (state variables also constitute the information stored in memory); a \textbf{policy} that takes the state variable and the memory's recurrent states as input to generate an action distribution; a \textbf{decoder}, which takes in the state variable, and predicts the value function as well as all current observations.

We now describe the model in detail by defining its parts  and the loss functions used to optimise it. Parameters given per task are defined in Table~\ref{table:parameters}.

\subsection{Encoder}
The encoder is composed of three sub-networks: the image encoder, the action encoder, and the reward encoder. These act independently on the different elements contained within the input set $o_t \equiv (I_t, a_{t-1}, r_{t-1})$, where $I_t$ is the current observed image, and $a_{t-1}$ and $r_{t-1}$ are the action and reward of previous time step. The outputs from these sub-networks are concatenated into a flat vector $e_t$. 

\subsubsection{Image Encoder}
The image encoder takes in image tensors $I_t$ of size $64 \times 64 \times 3$ (3 channel RGB). We then apply 6 ResNet~\cite{he2016deep} blocks with rectified linear activation functions. All blocks have 64 output channels and bottleneck channel sizes of 32. The strides for the 6 blocks are $(2, 1, 2, 1, 2, 1)$, resulting in 8-fold spatial down-sampling of the original image. Therefore, the ResNet module outputs tensors of size $8 \times 8 \times 64$. We do not use batch normalization~\cite{ioffe2015batch}, a pre-activation function on inputs, or a final activation function on the outputs. Finally, the output of the ResNet is flattened (into a $4,\!096$-element vector) and then propagated through one final linear layer that reduces the size to 500 dimensions, whereupon a $\tanh$ nonlinearity is applied.

\subsubsection{Action Encoder}
In all environments, the action from the previous time step is a one-hot binary vector $a_{t-1}$ (6-dimensional here) with $a_0 \equiv 0$. We use an identity encoder for the action one-hot.

\subsubsection{Reward Encoder}
The reward from the previous time step $r_{t-1}$ is also processed by an identity encoder.

\subsection{Decoder}
The decoder is composed of four sub-networks.
Three of these sub-networks are matched to the encoder sub-networks of image, previous action, and previous reward. The fourth sub-network decodes the value function.

\subsubsection{Image Decoder}
The image decoder has the same architecture as the encoder except the operations are reversed. In particular, all 2D convolutional layers are replaced with transposed convolutions~\cite{dumoulin2016guide}. Additionally, the last layer produces a number of output channels that parameterize the likelihood function used for the image reconstruction loss, described in more detail in Eq.~\ref{eq:output_loss}.

\subsubsection{Action and Reward Decoders}
The reward and action decoders are both linear layers from the state variable, $z_t$, to, respectively, a scalar dimension and the action cardinality.

\subsubsection{Value Function Predictor}
The value function predictor is a multi-layer perceptron (MLP) that takes in the concatenation of the state variable with the action distribution's logits, where, to ensure that the value function predictor learning does not modify the policy, we block the gradient (stop gradient) back through to the policy logits. The MLP has a single hidden layer of $200$ hidden units and a $\tanh$ activation function, which then projects via another linear layer to a 1-dimensional output. This function is a state-value function
$\hat{V}_t^\pi \equiv \hat{V}^\pi(z_t, \, \sg{\log \pi_t})$.

\subsection{Memory-Augmented RNN}
The RNN is primarily based on a simplification of the Differentiable Neural Computer (DNC)~\cite{graves2016hybrid}. It is composed of a deep LSTM and a slot-based external memory. The LSTM has recurrent state $(h_t, c_t)$ (output state and cells, respectively). The memory itself is a two-dimensional matrix $M_t$ of size $N \times W$, where $W$ is the same size as a state variable $z$. The memory at the beginning of each episode is initialised blank, namely $M_0 = 0$. We also carry the memory readouts $m_t \equiv [m_t^{(1)}, m_t^{(2)}, \dots, m_t^{(k)}]$, which is a list of $k$ vectors read from the memory $M_t$, as recurrent state. 

 At each time step, the following steps are taken sequentially:
 \begin{enumerate}
     \item Generate the state variable $z_t$ with $e_t$, $h_{t-1}$, and $m_{t-1}$ as input.
     \item Update the deep LSTM state with $h_t=\text{LSTM}(z_t, m_{t-1}, h_{t-1})$.
     \item Construct the read key and read from the external memory.
     \item Write the state variable $z_t$ to a new slot in the external memory.
 \end{enumerate}

\subsubsection{State Variable Generation}
The first step is to generate a state variable, $z_t$, combining both the new observation with the recurrent information. We take the encoded current observation $e_t$ concatenated with the recurrent information $h_{t-1}$ and $m_{t-1}$ as input through a single hidden-layer MLP with the hidden layer of size $2 \times W$ $\tanh$ units and output layer of size $W$.
\subsubsection{Deep LSTMs}
We use a deep LSTM~\cite{graves2013speech} of two hidden layers. Although the deep LSTM model has been described before, we describe it here for completeness. Denote the input to the network at time step $t$ as $x_t$. Within a layer $l$, there is a recurrent state $h^l_t$ and a ``cell'' state $c_t^l$, which are updated based on the following recursion (with $\sigma(x) \equiv (1 + \exp(-x))^{-1}$):
\begin{align*}
i^l_t &= \sigma\left(W_i^l [x_t, h^l_{t-1},h^{l-1}_{t}]  + b^l_i \right)\\
f^l_t &= \sigma\left(W_f^l [x_t, h^l_{t-1},h^{l-1}_{t}]  + b^l_f \right)\\
c^l_t &= f_t^l s_{t-1}^l + i_t^l \tanh \left(W_s^l [x_t, h^l_{t-1},h^{l-1}_{t}] + b^l_s  \right)\\
o^l_t &= \sigma\left(W_o^l [x_t, h^l_{t-1},h^{l-1}_{t}]  + b^l_o \right)\\
h^l_t &= o^l_t \tanh(c^l_t) 
\end{align*}
To produce a complete output $h_t$, we concatenate the output vectors from each layer: $h_t \equiv [h_t^1, h_t^2]$. These are passed out for downstream processing.

\subsubsection{LSTM Update}
At each time step $t$, the deep LSTM receives input $z_t$, which is then concatenated with the memory readouts at the previous time step $m_{t-1}$. The input to the LSTM is therefore $x_t = [z_t, m_{t-1}]$. The deep LSTM equations are applied, and the output $h_t$ is produced. 

\subsubsection{External Memory Reading}
\label{External Memory Reading}
A linear layer is applied to the LSTM's output $h_t$ to construct a memory interface vector $i_t$ of dimension $k \times (W + 1)$. The vector $i_t$ is then segmented into $k$ read keys $k_t^{(1)}, k_t^{(2)}, \dots, k_t^{(k)}$ of length $W$ and $k$ scalars ${sc}_t^{(1)}, \dots, {sc}_t^{(k)}$, which are passed through the function $\text{SoftPlus}(x) = \log(1 + \exp(x))$ to produce the scalars $\beta_t^{(1)}, \beta_t^{(2)} \dots, \beta_t^{(k)}$.

Memory reading is executed before memory writing. Reading is content-based. Reading proceeds by computing the cosine similarity between each read key $k_t^{(i)}$ and each memory row $j$: $c^{(ij)}_t = \cos(k_t^{(i)}, M_{t-1}[j, \cdot]) = \frac{k_t^{(i)} \cdot M_{t-1}[j, \cdot]}{|k_t^{(i)}| |M_{t-1}[j, \cdot]|}$. We then find indices $j_1^{(i)}, \dots, j_{\text{top}_K}^{(i)}$ corresponding to the $\text{top}_K$ largest values of $c^{(ij)}_t$ (over index $j$). 
Note that since unwritten rows of $M_{t-1}$ are equal to the zero vector, some of the chosen $j_1, \dots, j_{\text{top}_K}$ may correspond to rows of $M_{t-1}$ that are equal to the zero vector.

A weighting vector of length $N$ is then computed by setting: 
\begin{align*}
w_t^{(i)}[j] & = 
\begin{cases}
\frac{\displaystyle \exp (\beta_t^{(i)} c^{(ij)}_t)}{\displaystyle \sum_{j' \in \{ j_1^{(i)}, \dots, j_{\text{top}_K}^{(i)} \} } \exp (\beta_t^{(i)} c^{(ij')}_t)}, \, \text{for } j \in \big \{ j_1^{(i)}, \dots, j_{\text{top}_K}^{(i)} \big \} \\
0, \, \text{otherwise}.
\end{cases}    
\end{align*}
For each key, the readout from memory is $m_t^{(i)} = M_{t-1}^\top w_t^{(i)}$. The full memory readout is the concatenation across all read heads: $m_t \equiv [m_t^{(1)}, \dots, m_t^{(k)}]$.

\subsubsection{External Memory Writing}
Writing to memory occurs after reading, which we also define using weighting vectors. The write weighting $\ww_t$ has length $N$ and always appends information to the $t$-th row of the memory matrix at time $t$, i.e., $\ww_t[i] = \delta_{it}$ (using the Kronecker delta). The information we write to the memory is the state variable $z_t$. Thus, the memory update can be written as
\begin{align}
M_t & = M_{t-1} + \ww_t z_t^\top,   
\label{eq:mem_update}
\end{align}

\subsection{Policy}
The policy module receives $z_t$, $h_t$, and $m_t$ as inputs. The inputs are passed through a single hidden-layer MLP with 200 $\tanh$ units. This then projects to the logits of a multinomial softmax with the dimensionality of the action space. The action $a_t$ is sampled and executed in the environment.

\section{Loss Functions}
\label{sec:mbp_and_policy_loss}
We combine a policy gradient loss with reconstruction objectives for decoding observations. We also have a specific loss that regularizes the use of memory for TVT.

\subsection{Reconstruction Loss}
The reconstruction loss is the negative conditional log-likelihood of the observations and return, i.e.\ $- \log p(o_t, R_t | z_t)$, which is factorised into independent loss terms associated with each decoder sub-network and is conditioned on the state variable $z_t$. We use a multinomial softmax cross-entropy loss for the action, mean-squared error (Gaussian with fixed variance of 1) losses for the reward and the value function, and a Bernoulli cross-entropy loss for each pixel channel of the image. Thus, we have a negative conditional log-likelihood loss contribution at each time step of
\begin{align}
- \log p(o_t, R_t | z_t) \equiv & \alpha_\text{image} \mathcal{L}_\text{image}
+ \alpha_\text{value} \mathcal{L}_\text{value} + \alpha_\text{rew} \mathcal{L}_\text{rew}  + \alpha_\text{act} \mathcal{L}_\text{act},
\label{eq:output_loss}
\end{align}
where 
\begin{align}
\mathcal{L}_\text{image} & = \sum_{w=1,h=1,c=1}^{|W|,|H|,|C|} \bigg[ I_t[w,h,c] \log \hat{I}_t[w,h,c] + (1 - I_t[w,h,c]) \log (1 - \hat{I}_t[w,h,c] ) \bigg] \nonumber, \\
\mathcal{L}_\text{value} & = \frac{1}{2} \bigg [ || R_t - \hat{V}^\pi(z_t, \sg{\log \pi_t}) ||^2 \bigg ], \nonumber \\
 \mathcal{L}_\text{rew} & = \frac{1}{2} || r_{t-1} - \hat{r}_{t-1} ||^2, \nonumber \\
\mathcal{L}_\text{act} & = \sum_{i=1}^{|A|} \bigg [a_{t-1}[i] \log ( \hat{a}_{t-1}[i]) + (1-a_{t-1}[i]) \log (1 - \hat{a}_{t-1}[i]) \bigg ] \nonumber, 
 \bigg ] \nonumber.
\end{align}

On all but the standard RL control experiment tasks, we constructed the target return value as $R_t = r_t + \gamma r_{t+1} + \gamma^2 r_{t+2} + \dots + \gamma^{T-t} r_T$. For the standard RL control experiment tasks with episodes of length $T$, we use ``truncation windows''~\cite{mnih2016asynchronous} in which the time axis is subdivided into segments of length $\tau_\text{window}$. We can consider full gradient as a truncated gradient with $\tau_\text{window} = T$. If the window around time index $t$ ends at time index $k$, the return within the window is 
\begin{align}
R_t : =
\begin{cases}
r_t + \gamma r_{t+1} + \gamma^2 r_{t+2} + \dots + \gamma^{k - t + 1} \hat{V}^\pi_\nu(z_{k + 1}, \log \pi_{k + 1}), & \text{ if } k < T,\\
r_t + \gamma r_{t+1} + \gamma^2 r_{t+2} + \dots + \gamma^{T - t} r_T, & \text{ if } T \leq k.
\end{cases}
\label{eq:bootstrap}
\end{align}

As a measure to balance the magnitude of the gradients from different reconstruction losses, the image reconstruction loss is divided by the number of pixel-channels $|W| \times |H| \times |C|$.

\subsection{Policy Gradient}
We use discount and bootstrapping parameters $\gamma$ and $\lambda$, respectively, as part of the policy advantage calculation given by the Generalised Advantage Estimation (GAE) algorithm \cite{schulman2015high}. Defining $\delta_t \equiv r_t + \gamma \hat{V}^\pi(z_{t+1}, \log \pi_{t+1}) - \hat{V}^\pi(z_{t}, \log \pi_{t})$, Generalised Advantage Estimation makes an update of the form:

\begin{equation}
\Delta \theta \propto \sum_{t=k \tau_\text{window}}^{(k+1) \tau_\text{window}} \sum_{t'=t}^{(k+1) \tau_\text{window}} (\gamma \lambda)^{t'-t} \delta_{t'} \nabla_\theta \log \pi_\theta(a_t | h_t). \label{eq:policy_gradient_detailed}
\end{equation}
There is an additional loss term that increases the entropy of the policy's action distribution. This and pseudocode for all of RMA's updates are provided in Algorithm~\ref{alg:rma}.

\subsection{Temporal Value Transport Specific Loss}
We include an additional regularization term described in Section \ref{sec:read_reg}.

\section{Comparison Models}
We introduce two comparison models: the LSTM+Mem Agent and the LSTM Agent.

\subsection{LSTM+Mem Agent}
The LSTM+Mem Agent is similar to the RMA. The key difference is that it has no reconstruction decoders and losses. The value function is produced by a one hidden-layer MLP with 200 hidden units: $\hat{V}(z_t, \sg{\log \pi_t})$.

\subsection{LSTM Agent}
The LSTM Agent additionally has no external memory system and is essentially the same design as the A3C agent~\cite{mnih2016asynchronous}. We have retrofitted the model to share the same encoder networks as the RMA, acting on input observations to produce the same vector $e_t$. This is then passed as input to a deep 2-layer LSTM that is the same as the one in RMA. The LSTM has two output ``heads'', which are both one hidden-layer MLPs with 200 hidden units: one for the policy distribution $\pi(a_t | z_t, h_t)$ and one for the the value function prediction $\hat{V}(z_t, h_t, \sg{\log \pi_t})$. As for our other agents, the policy head is trained using Eq.~\ref{eq:policy_gradient_detailed}.

\section{Implementation and Optimisation}
For optimisation, we used truncated backpropagation through time~\cite{sutskever2013training}. We ran 384 parallel worker threads that each ran an episode on an environment and calculated gradients for learning. Each gradient was calculated after one truncation window, $\tau_\text{window}$. For all main paper experiments other than the standard RL control experiments, $\tau_\text{window}=T$, the length of the episode.

The gradient computed by each worker was sent to a ``parameter server'' that asynchronously ran an optimisation step with each incoming gradient. We optimise the model using ADAM optimisers~\cite{kingma2014adam} with $\beta_1=0.9$ and $\beta_2=0.999$. 

The pseudocode for each RMA worker is presented in Algorithm~\ref{alg:rma}. 

\begin{algorithm}
\caption{RMA Worker Pseudocode}
\label{alg:rma}
\begin{algorithmic}
\STATE // Assume global shared model parameter vectors $\theta$ and counter $T := 0$
\STATE // Assume thread-specific parameter vectors $\theta'$
\STATE // Assume discount factor $\gamma \in (0,1]$ and bootstrapping parameter $\lambda \in [0,1]$
\STATE Initialize thread step counter $t := 1$
\REPEAT
\STATE Synchronize thread-specific parameters $\theta' := \theta$
\STATE Zero model's memory \& recurrent state if new episode begins
\STATE $t_\text{start} := t$
\REPEAT
\STATE $e_t = \text{Encode}(o_t)$

\STATE $z_t = \text{StateVariableMLP}(e_t, h_{t-1}, m_{t-1})$

\STATE $h_t, m_t = \text{RNN}(z_t, h_{t-1}, m_{t-1})$  // (Memory-augmented RNN)
\STATE Update memory $M_t = \text{Write}(M_{t-1}, z_t)$

\STATE Policy distribution $\pi_t = \pi(a_t | z_t, h_t, m_t)$
\STATE Sample $a_t \sim \pi_t$

\STATE $\hat{V}_t, o_t^{r} = \text{Decode}(z_t, \sg{\log \pi_{t}})$
\STATE Apply $a_t$ to environment and receive reward $r_t$ and observation $o_{t+1}$
\STATE $t := t + 1; T := T + 1$
\UNTIL{environment termination or $t-t_\text{start} == \tau_\text{window}$}

\STATE If not terminated, run additional step to compute $\hat{V}_\nu(z_{t+1}, \log \pi_{t+1})$
\STATE and set $R_{t+1} := \hat{V}(z_{t+1}, \log \pi_{t+1})$ // (but don't increment counters)

\STATE \textbf{(Optional) Apply Temporal Value Transport  (Alg.~\ref{alg:TVT})}

\STATE Reset performance accumulators $\mathcal{A} := 0; \mathcal{L} := 0; \mathcal{H} := 0$
     \FOR{$k$ from $t$ down to $t_\text{start}$}
        \STATE $\gamma_t :=
            \begin{cases}
            0, \text{ if } k \text{ is environment termination} \\
            \gamma, \text{ otherwise } \\
            \end{cases}$ \\
        \STATE $R_k := r_k + \gamma_t R_{k+1}$      
        \STATE $\delta_k := r_k + \gamma_t \hat{V}(z_{k+1}, \log \pi_{k+1}) - \hat{V}(z_{k}, \log \pi_{k})$ 
        \STATE $A_k := \delta_k + (\gamma \lambda) A_{k+1}$
        \STATE $\mathcal{A} := \mathcal{A} + A_k \log \pi_k[a_k]$
        \STATE $\mathcal{H} := \mathcal{H} - \alpha_\text{entropy} \sum_i \pi_k[i] \log \pi_k[i]$ $\text{ // (Entropy loss) }$
        \STATE $\mathcal{L} := \mathcal{L} + \mathcal{L}_k$ (Eq.~\ref{eq:output_loss})
\ENDFOR
\STATE $d \theta' := \nabla_{\theta'} (\mathcal{A} + \mathcal{H} + \mathcal{L})$
\STATE Asynchronously update via gradient ascent $\theta$ using $d \theta'$
\UNTIL{$T > T_\text{max}$}
\end{algorithmic}
\end{algorithm}

For all experiments, we used the open source package Sonnet -- available at \url{https://github.com/deepmind/sonnet} -- and applied its defaults to initialise network parameters.

\section{Temporal Value Transport}
Temporal Value Transport works in two stages. First, we identify significant memory read events, which become splice events. Second, we transport the value predictions made at those read events back to the time points being read from, where they modify the rewards and therefore the RL updates.

\subsection{Splice Events}
At time $t'$, the read strengths $\beta_{t'}^{(i)}$ are calculated as described in \ref{External Memory Reading}. To exclude sending back value to events in the near past, for time points $t'$ where $t' - \arg \max_t w_{t'}[t] < 1/ (1 - \gamma)$, we reset $\beta_{t'}^{(i)} : = 0$ for the remainder of the computation. We then identify splice events by first finding all time windows $[t_{-}' , t_{+}']$ where $\beta_{t'}^{(i)} \geq \beta_\text{threshold}$ for $t' \in [t_{-}', t_{+}']$ but $\beta_{t'}^{(i)} < \beta_\text{threshold}$ for $t'=t_{-}' - 1$ and $t'=t_{+}' + 1$. 

We then set $t_{\max}$ to be the $\arg \max$ over $t'$ of $\beta_{t'}^{(i)}$ in the period for the included points. 
 
\subsection{Reward Modification}
For each $t_\text{max}$ above, we modify the reward of all time points $t$ occurred more than $1/(1-\gamma)$ steps beforehand: 
\begin{align}
r_t \to \begin{cases}
r_t + \alpha w_{t_\text{max}}^{(i)}[t] \hat{V}_{t_\text{max} + 1}, \,  \text{if } t > t_\text{max} - 1 / (1 - \gamma), \\  
r_t, \, \text{otherwise.}\\
\end{cases}
\end{align}
We send back $\hat{V}_{t_\text{max} + 1}$ because that is the first value function prediction that incorporates information from the read at time $t_\text{max}$. Additionally, for multiple read processes $i$, the process is the same, with independent, additive changes to the reward at any time step. Pseudocode for Temporal Value Transport with multiple read processes is provided in Algorithm~\ref{alg:TVT}.

\subsection{Reading Regularization}
\label{sec:read_reg}
To prevent the TVT mechanism from being triggered extraneously, we impose a small regularization cost whenever a read strength is above threshold.
\begin{align}
\mathcal{L}_\text{read-regularization} & = 5 \times 10^{-6} \times \sum_{i=1}^k  \max(\beta_t^{(i)} - \beta_\text{threshold}, 0). 
\end{align}
This is added to the other loss terms.

\begin{algorithm}[ht]
\caption{Temporal Value Transport for Multiple Reads}
\begin{algorithmic}
    \STATE \textbf{input:} $\{r_t\}_{t \in [1,T]}$, $\{\hat{V}_t\}_{t \in [1,T]}$, 
    read strengths $\{\beta_t^{(i)}\}_{t \in [1,T],i \in [1,k]}$, 
    read weights $\{w_t^{(i)}\}_{t \in [1,T],i \in [1,k]}$
    \FOR{$i \in [1,k]$}
        \FOR{$t' \in [1,T]$} 
            \IF{$t' - \arg \max_t w_{t'}^{(i)}[t] < 1 / (1 - \gamma)$} 
                \STATE $\beta_{t'}^{(i)} : = 0$    
            \ENDIF
        \ENDFOR

        \STATE $\text{splices} : = []$ 
        \FOR{\textbf{each} crossing of read strength $\beta_t^{(i)}$ above $\beta_\text{threshold}$}
            \STATE $t_\text{max} : = \arg \max_t \{\beta_t^{(i)} | t \in \text{crossing window} \}$
            \STATE Append $t_\text{max}$ to splices 
        \ENDFOR
        \FOR{$t$ in 1 to T}       
            \FOR{$t'$ in splices}
                \IF{$t < t' - 1/(1-\gamma)$} 
                    \STATE $r_t := r_t + \alpha w_{t'}^{(i)}[t] \hat{V}_{t'+1}$ 
                \ENDIF
            \ENDFOR
        \ENDFOR
    \ENDFOR

    \RETURN $\{r_t\}_{t \in [1,T]}$
\end{algorithmic}
\label{alg:TVT}
\end{algorithm}
\break

\section{Signal-to-Noise Ratio Analysis} 
\subsection{Undiscounted Case}
As in the article text, we refer to phases 1-3 as P1-P3.
We define the signal as the squared norm of the expected policy change in P1 induced by the policy gradient. To be precise, let $\Delta \theta : = \sum_{t \in P1} \nabla_\theta \log \pi(a_t | h_t) R_{t}$. Further, in the following assume that the returns are baseline-subtracted, i.e.\ $R_t \to R_t - \E_\pi[R_t]$. We define the signal as 
\begin{align}
\text{Signal} & := \| \E_\pi[\Delta \theta] \|^2 \nonumber \\
& = \bigg \| \E_\pi \bigg[\sum_{t \in P1} \nabla_\theta \log \pi(a_t | h_t) \sum_{t' \geq t} r_{t'} \bigg] \bigg \|^2. \nonumber   
\end{align}
We define the noise as the trace of the variance of the policy gradient
\begin{align}
\text{Noise} & := \Tr \big ( \text{Var}_\pi [\Delta \theta]) \nonumber \\
& = \mathbb{E}_\pi \bigg [\bigg \| \sum_{t \in P1} \nabla_\theta \log \pi(a_t | h_t) R_t - \E_\pi[\Delta \theta] \bigg \|^2  \bigg ]. \nonumber
\end{align}
Recall that $r_t \equiv 0$ for $t \in \text{P1}$. Further, P1 and P2 are approximately independent as P2 is a distractor phase whose initial state is unmodified by activity in P1. The only dependence is given by the agent's internal state and parameters, but we assume for these problems it is a weak dependence, which we ignore for present calculations. In this case,
\begin{align}
\mathbb{E}_\pi \bigg[\sum_{t \in P1} \nabla_\theta \log \pi(a_t | h_t) \sum_{t' \geq t} r_{t'}  \bigg ] & = \mathbb{E}_\pi \bigg[\sum_{t \in P1} \nabla_\theta \log \pi(a_t | h_t) \bigg[\sum_{t' \in P2} r_{t'} + \sum_{t' \in P3} r_{t'}\bigg] \bigg ] \nonumber \\
& \approx \mathbb{E}_\pi \bigg[\sum_{t \in P1} \nabla_\theta \log \pi(a_t | h_t) \sum_{t' \in P3} r_{t'} \bigg ]. 
\end{align}
Based on these considerations, the signal term is easy to calculate: 
\begin{align}
\text{Signal} & \approx \| \mathbb{E}_\pi [\Delta \theta | \text{no P2}] \|^2 \nonumber \\
& = \bigg \| \mathbb{E}_\pi \bigg [\sum_{t \in P1} \nabla_\theta \log \pi(a_t | h_t) \sum_{t' \in P3} r_{t'}  \bigg ] \bigg \|^2. \nonumber
\end{align}
Define $g_\theta : = \sum_{t \in P1} \nabla_\theta \log \pi(a_t | h_t)$. With this, the noise term becomes
\begin{align}
\text{Noise} & = \mathbb{E}_\pi \bigg [\bigg \| \sum_{t \in P1} \nabla_\theta \log \pi(a_t | h_t) \sum_{t' \geq t} r_{t'} - \E_\pi[\Delta \theta] \bigg \|^2  \bigg ] \nonumber \\
& = \mathbb{E}_\pi \bigg [\bigg \| g_\theta \bigg [\sum_{t' \in P2} r_{t'} + \sum_{t' \in P3} r_{t'} \bigg] - \E_\pi[\Delta \theta] \bigg \|^2  \bigg ] \nonumber \\
& = \mathbb{E}_\pi \bigg [\bigg \| g_\theta\sum_{t' \in P2} r_{t'} + \bigg ( g_\theta \sum_{t' \in P3} r_{t'}  - \E_\pi[\Delta \theta] \bigg ) \bigg] \bigg \|^2  \bigg ] \nonumber \\
& \approx \mathbb{E}_\pi \bigg [\bigg \| g_\theta \sum_{t' \in P2} r_{t'} \bigg \|^2  \bigg ] + \Tr \big ( \text{Var}_\pi [\Delta \theta | \text{no P2}] \big ), \nonumber
\end{align}
where $\Tr \big ( \text{Var}_\pi [\Delta \theta | \text{no P2}] \big )$ is the variance in the policy gradient due to P1 and P3 without a P2 distractor phase. (The approximate equality represents that the memory state of the system is altered by the P2 experience, but we neglect this dependence.) 
We make the assumption that performance in P2 is independent of activity in P1, which is approximately the case in the distractor task we present in the main text. With this assumption, the first term above becomes
\begin{align}
\mathbb{E}_\pi \bigg [\bigg \| \sum_{t \in P1} \nabla_\theta \log \pi(a_t | h_t) \sum_{t' \in P2} r_{t'} \bigg \|^2  \bigg ] & = \text{Var}_\pi \bigg [ \sum_{t' \in P2} r_{t'} \bigg ] \times \mathbb{E}_\pi \bigg [\bigg \| \sum_{t \in P1} \nabla_\theta \log \pi(a_t | h_t) \bigg \|^2 \bigg ] \nonumber \\
& = \text{Var}_\pi \bigg [ \sum_{t' \in P2} r_{t'} \bigg ] \times \mathbb{E}_\pi \bigg [\bigg \| \sum_{t \in P1} \nabla_\theta \log \pi(a_t | h_t) \bigg \|^2 \nonumber \\ & \, \, \, \, \, \, - \bigg \| \underbrace{\mathbb{E}_\pi \bigg [\sum_{t \in P1} \nabla_\theta \log \pi(a_t | h_t) \bigg ]}_{=0} \bigg \|^2 \bigg ] \nonumber \\
& = \text{Var}_\pi \bigg [ \sum_{t' \in P2} r_{t'} \bigg ] \times \Tr \bigg ( \text{Var}_\pi \bigg [\sum_{t \in P1} \nabla_\theta \log \pi(a_t | h_t) \bigg ] \bigg ) \nonumber.
\end{align}
Thus, the SNR (Signal / Noise) is approximately
\begin{align}
\mathrm{SNR} & \approx \frac{\displaystyle \bigg \| \mathbb{E}_\pi \bigg [\sum_{t \in P1} \nabla_\theta \log \pi(a_t | h_t) \sum_{t' \in P3} r_{t'}  \bigg ] \bigg \|^2}{\displaystyle \text{Var}_\pi \bigg [ \sum_{t' \in P2} r_{t'} \bigg ] \times \Tr \bigg ( \text{Var}_\pi \bigg [\sum_{t \in P1} \nabla_\theta \log \pi(a_t | h_t) \bigg ] \bigg ) + \Tr \big ( \text{Var}_\pi [\Delta \theta | \text{no P2}] \big )} \nonumber.
\end{align}
In the limit of large P2 reward variance, we have
\begin{align}
\mathrm{SNR} & \approx \frac{\displaystyle \bigg \| \mathbb{E}_\pi \bigg [\sum_{t \in P1} \nabla_\theta \log \pi(a_t | h_t) \sum_{t' \in P3} r_{t'}  \bigg ] \bigg \|^2}{\displaystyle \text{Var}_\pi \bigg [ \sum_{t' \in P2} r_{t'} \bigg ] \times \Tr \bigg ( \text{Var}_\pi \bigg [\sum_{t \in P1} \nabla_\theta \log \pi(a_t | h_t) \bigg ] \bigg )} \nonumber.
\end{align}
The reward variance in P2, $\text{Var}_\pi \big [ \sum_{t' \in P2} r_{t'} \big ]$, reduces the policy gradient SNR, and low SNR is known to impact the convergence of stochastic gradient optimization negatively \cite{roberts2009signal}. Of course, averaging $S$ independent episodes increases the SNR correspondingly to $S \times \mathrm{SNR}$, but the approach of averaging over an increasing number of samples is not universally possible and only defers the difficulty: there is always a level of reward variance in the distractor phase that matches or overwhelms the variance reduction achieved by averaging. 

\section{Tasks}

All tasks were implemented in DeepMind Lab (DM Lab)~\cite{beattie2016deepmind}.
DM Lab has a standardized environment map unit length: all sizes given below are in these units.

\subsection{Observation and Action Repeats}
For all DM Lab experiments, agents processed 15 frames per second. The environment itself produced 60 frames per second, but we propagated only the first observation of each packet of four to the agents. Rewards accumulated over each packet were summed together and associated to the first, undropped frame. Similarly, the agents chose one action at the beginning of this packet of four frames: this action was applied four times in a row. We define the number of ``Agent Steps'' as the number of independent actions sampled by the agent: that means one agent step per packet of four frames.

\subsection{Action Sets}

We used a consistent action set for all experiments except for the Arbitrary Visuomotor Mapping task. For all other tasks, we used a set of six actions: \emph{move forward}, \emph{move backward}, \emph{rotate left with rotation rate of 30} (mapping to an angular acceleration parameter in DM Lab), \emph{rotate right with rotation rate of 30}, \emph{move forward and turn left}, \emph{move forward and turn right}. For the Arbitrary Visuomotor Mapping, we did not need to move relative to the screen, but we instead needed to move the viewing angle of the agent. We thus used four actions: \emph{look up}, \emph{look down}, \emph{look left}, \emph{look right} (with rotation rate parameter of 10).

\subsection{Themes}

DM Lab maps use \emph{texture sets} to determine the floor and wall textures. We use a combination of four different texture sets in our tasks: \emph{Pacman}, \emph{Tetris}, \emph{Tron} and \emph{Minesweeper}. DM Lab texture sets can take on various colours but we use the default colours for each set, which are:
Pacman: blue floors and red walls.
Tetris: blue floor and yellow walls.
Tron: yellow floor and green walls.
Minesweeper: blue floor and green walls.
Examples of how these sets appear can be seen in various figures in the main text.

\subsection{Task Phases}
Episodes for the tasks with delay intervals are broken up into multiple phases. Phases do not repeat within an episode. Generally, the tasks contain three phases (P1-P3), with a middle phase.

We used a standardized P2 distractor phase task: the map is an $11 \times 11$ open square (Figure~1b second column). The agent spawns (appears) adjacent to the middle of one side of the square, facing the middle. An apple is randomly spawned independently at each unit of the map with probability $0.3$, except for the square in which the agent spawns. Each apple gives a reward $r_\text{apple}$ of 5 when collected and disappears after collection. The agent remains in this phase for 30 seconds. (This length was varied in some experiments.) The map uses the Tetris texture set unless mentioned otherwise.

\subsection{Cue Images}
In several tasks, we use \emph{cue images} to provide visual feedback to the agent, e.g., indicating that an object has been picked up. These cue images are colored rectangles that overlay the input image, covering the majority of the top half of the image. An example of a red cue image is shown in Supplementary Figure~\ref{fig:suppfig10}a, third panel. These cues are shown for 1 second once activated, regardless of a transition to a new phase that may occur during display.

\subsection{Primary Tasks}
\subsubsection{Passive Visual Match}

In each episode of Passive Visual Match, four distinct colors are randomly chosen from a fixed set of 16 colors. One of these is selected as the \emph{target color} and the remaining three are \emph{distractor colors}. Four squares are generated with these colors, each the size of one wall unit. The three phases in each episode are:

\begin{enumerate}
\item The map is a $1 \times 3$ corridor with a target color square covering the wall unit at one end. The agent spawns facing the square from the other end of the corridor (Figure~1b first column). There are no rewards in this phase. The agent remains in this phase for 5 seconds. The map uses the Pacman texture set.
\item The standard distractor phase described above.
\item The map is a $4 \times 7$ rectangle with the four color squares (the target color and three distractor colors) on one of the longer sides, with a unit gap between each square. The ordering of the four colors is randomly chosen. There is an additional single unit square placed in the middle of the opposite side, in which the agent spawns, facing the color squares. In front of each color square is a groundpad (Figure~1b last two columns). When the agent touches one of these pads, a reward of 10 points is given if it is the pad in front of the target painting and a reward of 1 is given for any other pad. The episode then ends. If the agent does not touch a pad within 5 seconds then no reward is given for this phase and the episode ends. The map uses the Tron texture set.
\end{enumerate}

\subsubsection{Active Visual Match}
\label{ActiveVisualMatch}
Active Visual Match is the same as Passive Visual Match, except that the map in phase 1 is now larger and the position of the target image in phase 1 is randomized. The phase 1 map consists of two $3 \times 3$ open squares connected by a $1 \times 1$ corridor that joins each square in the middle of one side (Figure~2a first two columns). The agent spawns in the center of one of the two squares, facing the middle of one the walls adjacent to the wall with the opening to the corridor. The target color square is placed randomly over one of any of the wall units on the map.

\subsubsection{Key-to-Door}

The three phases of Key-to-Door are:

\begin{enumerate}
\item The map is identical to the map in phase 1 of Active Visual Match. The agent spawns in the corner of one the squares that is furthest from the opening to the corridor, facing into the square but not towards the opening. A key is placed randomly within the map (not at the spawn point) and if the agent touches the key it disappears and a black cue image is shown (Figure~4a first two columns). As in the Visual Match tasks, there are no rewards in this phase, and the phase lasts for 5 seconds. The map uses the Pacman texture set.
\item The standard distractor phase.
\item The map is a $1 \times 3$ corridor with a locked door in the middle of the corridor. The agent spawns at one end of the corridor, facing the door. At the end of the corridor on the other side of the door is a goal object (Figure~4a fourth column). If the agent touched the key in phase one, the door can be opened by walking into it, and then if the agent walks into the goal object a reward of 10 points is given. Otherwise, no reward is given. The map uses the Tron texture set.
\end{enumerate}

\subsubsection{Key-to-Door-to-Match}
This task combines elements of Key-to-Door with Active Visual Match. One target color and three distractor colors are chosen in the same way as for the Visual Match tasks. In contrast to the standard task setup, there are five phases per episode:

\begin{enumerate}
\item This phase is the same as phase 1 of Key-to-Door but with a different map. The map is a $3 \times 4$ open rectangle with an additional $1 \times 1$ square attached at one corner, with the opening on the longer of the two walls. The agent spawns in the additional $1 \times 1$ square, facing into the rectangle (Figure~5a first column). The map uses the Minesweeper texture set.
\item The standard distractor phase except that the phase lasts for only 15 seconds instead of 30 seconds.
\item The map is the same as in phase 3 of Key-to-Door. Instead of a goal object behind the locked door, the target color square covers the wall at the far end of the corridor (Figure~5a third column). There is no reward in this phase, and it lasts for 5 seconds. The map uses the Pacman texture set.
\item The standard distractor phase except that the phase lasts for only 15 seconds instead of 30 seconds.
\item The final phase is the same as phase 3 in the Visual Match tasks.
\end{enumerate}

\subsubsection{Two Negative Keys}

The three phases of Two Negative Keys are:

\begin{enumerate}
\item The map is a $3 \times 4$ open rectangle. The agent spawns in the middle of one of the shorter walls, facing into the rectangle. One red key is placed in a corner opposite the agent, and one blue key is placed in the other corner opposite the agent. Which corner has the red key and which the blue key is randomized per episode. If the agent touches either of the keys, a red or blue cue image is shown according to which key the agent touched (Supplementary Figure~\ref{fig:suppfig10} first three columns). After one key is touched, it disappears, and nothing happens if the agent goes on to touch the remaining key (i.e., no cue is displayed and the key remains in the map). The phase lasts for 5 seconds, and there are no rewards; if the agent does not touch any key during this period, at the end of the phase a black cue image is shown. The map uses the Tron texture set.
\item The standard distractor phase except with the Tetris texture set.
\item The layout is the same as in phase 3 of the Key-to-Door task. If the agent has picked up either of the keys then the door will open when touched, and the agent can collect the goal object, at which point it will spawn back into the map from phase 2 but with all remaining apples removed. This phase lasts for only 2 seconds in total; when it ends, a reward of -20 is given if the agent did not collect the goal object; a reward of -10 is given if the agent collected the goal object after touching the blue key; and a reward of -1 is given if the agent collected the goal object after touching the red key. The map uses the Tron texture set.
\end{enumerate}

\subsubsection{Latent Information Acquisition}
In each episode, three objects are randomly generated using the DM Lab object generation utilities. Color and type of object is randomized. Each object is independently randomly assigned to be a \emph{good} or a \emph{bad} object.

\begin{enumerate}
\item The map is a $3 \times 5$ rectangle. The agent spawns in one corner facing outwards along one of the shorter walls. The three objects are positioned randomly among five points as displayed in Figure~6c in the main text (Figure~6a first four columns). If an agent touches one of the good objects, it disappears, and a green cue image is shown. If an agent touches one of the bad objects, it disappears, and a red cue image is shown. This phase lasts for 5 seconds, and there are no rewards. The map uses the Tron texture set. The image cues shown in this phase are only shown for 0.25 seconds so that the cues do not interfere with continuation of the P1 activity (in all other tasks they are shown for 1 second).
\item The standard distractor phase except with the Tetris texture set.
\item The map, spawn point, and possible object locations are the same as in phase 1. The objects are the same, but their positions are randomly chosen again. If the agent touches a good object it disappears, and a reward of 20 is given. If the agent touches a bad object it disappears and a reward of -10 is given. This phase lasts for 5 seconds. The map uses the Tron texture set.
\end{enumerate}

\subsection{Distractor Phase Modifications}
\label{sec:distractormods}
In order to analyze the effect of increasing variance of distractor reward on agent learning, we created variants of the distractor phase where this reward variance could be easily controlled. Since the distractor phase is standardized, any of these modifications can be used in any of those tasks.

\subsection{Zero Apple Reward}
\label{sec:zeroapple}

The reward given for apples in the distractor phase is zero. Even though the apples give zero reward, they still disappear when touched by the agent.

\subsection{Fixed Number of Apples}
\label{sec:fixednumapples}

The reward given for apples remains at 5. Instead of the 120 free squares of the map independently spawning an apple with probability 0.3, we fix the number of apples to be $120 \times 0.3 = 36$ and distribute them randomly among the 120 available map units. Under an optimal policy where all apples are collected, this has the same mean reward as the standard distractor phase but with no variance.

\subsection{Variable Apple Reward}
\label{sec:variableapple}

The reward $r_\text{apple}$ given for apples in the distractor phase can be modified (to a positive integer value), but with probability $1 - 1/r_\text{apple}$ each apple independently gives zero reward instead of $r_\text{apple}$. Any apple touched by the agent still disappears.

This implies that the optimal policy and expected return under the optimal policy is constant, but variance of the returns increases with $r_\text{apple}$. Since there are 120 possible positions for apples in the distractor phase, and apples independently appear in each of these positions with probability $0.3$, the variance of undiscounted returns in P2, assuming all apples are collected, is 
\begin{align}
120 \times \bigg [\bigg(0.3 \times \frac{1}{r_\text{apple}}\bigg) \times r_\text{apple}^2 - (0.3 \times 1)^2 \bigg ] = 36 \times (r_\text{apple} - 0.3).
\end{align}


\subsection{Control Tasks}

Control tasks are taken from the DM Lab 30 task set~\cite{beattie2016deepmind}. The tasks we include had a memory access component to performance. We provide only brief descriptions here since these tasks are part of the open source release of DM Lab available at \url{https://github.com/deepmind/lab/tree/master/game\_scripts/levels/contributed/dmlab30}. 

\subsubsection{Explore Goal Locations Small}

This task requires agents to find the goal object as fast as possible. Within episodes, when the goal object is found the agent respawns and the goal appears again in the same location. The goal location, level layout, and theme are randomized per episode. The agent spawn location is randomized per respawn.

\subsubsection{Natlab Varying Map Randomized}

The agent must collect mushrooms within a naturalistic terrain environment to maximise score. The mushrooms do not regrow. The map is randomly generated and of intermediate size. The topographical variation, and number, position, orientation and sizes of shrubs, cacti and rocks are all randomized. Locations of mushrooms are randomized. The time of day is randomized (day, dawn, night). The spawn location is randomized for each episode.

\subsubsection{Psychlab Arbitrary Visuomotor Mapping}
This is a task in the Psychlab framework\cite{leibo2018psychlab} where the agent is shown images from a visual memory capacity experiment dataset\cite{brady2008visual} but in an experimental protocol known as arbitrary visuomotor mapping. The agent is shown consecutive images that are associated to particular cardinal directions. The agent is rewarded if it can remember the direction to move its fixation cross for each image. The images are drawn from a set of roughly 2,500 possible images, and the specific associations are randomly generated per episode.

\subsection{Task Specific Parameters}
\label{sec:task_parameters}
Across models the same parameters were used for the TVT, RMA, LSTM+Mem, and LSTM agents except for $\gamma$, which for the TVT model was always $0.96$ and was varied as expressed in the figure legends for the other models. Learning rate was varied only for the learning rate analysis in Section~\ref{sec:lranalysis}.

Across tasks, we used the parameters shown in Table~\ref{table:parameters} with a few exceptions:
\begin{itemize}
\item For all the control tasks, we used $\alpha_\text{image}=1$ instead of 20.
\item For all the control tasks, we used $\tau_\text{window}=200$ instead of using the full episode.
\item For the Two Negative Keys task, we used $\alpha_\text{entropy}=0.05$ instead of $0.01$.
\end{itemize}

\begin{table}
\centering
 \begin{tabular}{||c | c ||} 
 \hline
 Parameter & Value  \\ [0.5ex] 
 \hline\hline
$\eta$ & $5 \times 10^{-6}$ \\
$\gamma$ & various \\
$\lambda$ & $= \gamma$ \\ 
$\alpha_\text{image}$ & 20 \\ 
$\alpha_\text{rew}$ & 1 \\ 
$\alpha_\text{value}$ & $0.4$ \\ 
$\alpha_\text{act}$ & 1 \\ 
$\alpha_\text{entropy}$ & 0.01 \\ 
$\tau_\text{window}$ & Number of steps in episode \\ 
$N$ & Number of steps in episode \\
$W$ & 200 \\ 
$k$ & 3 \\
$\text{top}_K$ & 50 \\
$\beta_\text{threshold}$ & 2 \\ [1ex] 
 \hline
 \end{tabular}
 \caption{\footnotesize{Parameters used across tasks (not all parameters apply to all models).}}
 \label{table:parameters}
\end{table}

\section{Task Analyses}

\subsection{Variance Analysis}

For Active Visual Match and Key-to-Door tasks, we performed analysis of the effect of distractor phase reward variance on the performance of the agents. To do this we used the same tasks but with modified distractor phases as described in Section~\ref{sec:distractormods}.

\subsection{Active Visual Match}
\label{sec:imagevar}

Supplementary Figure~\ref{fig:suppfig5} shows learning curves for $r_\text{apple}=0$ (see Section~\ref{sec:zeroapple}) and $r_\text{apple}=1$ (see section~\ref{sec:variableapple}). When $r_\text{apple}=1$, all apples give reward. Learning for the RMA was already significantly disrupted when $r_\text{apple}=1$, so for Active Visual Match we do not report higher variance examples.

\subsection{Key-to-Door}
\label{sec:ktdvar}

Figure~4c shows learning curves with apple reward $r_\text{apple}$ set to 1, 3, 6, and 10, which gives variances of total P2 reward as 25, 100, 196, and 361, respectively, (see section~\ref{sec:variableapple}). Note that episode scores for these tasks show that all apples are usually collected in P2 at policy convergence.

Note that the mean distractor phase return in the previous analysis is much less than the mean return in the standard distractor phase. Another way of looking at the effect of variance in the distractor phase whilst including the full mean return is shown in Supplementary Figure~\ref{fig:suppfig8}, which has three curves: one for zero apple reward (see~\ref{sec:zeroapple}), one for a fixed number of apples (see~\ref{sec:fixednumapples} and one for the full level (which has a variable number of apples per episode but the same expected return as the fixed number of apples case). From the figure, it can be seen that introducing large rewards slows learning in phase 1 due to the variance whilst the agent has to learn the policy to collect all the apples, but that the disruption to learning is much more significant when the number of apples continues to be variable even after the agent has learnt the apple collection policy.

\subsection{Return Prediction Saliency}
\label{sec:saliency}

To generate Figure 4e in the main text, a sequence of actions and observations for a single episode of Key-to-Door was recorded from a TVT agent trained on that level. We show two time steps where the key was visible. 
We calculated gradients $\partial \hat{V}_t / \partial I^{w,h,c}_t$ of the agent's value predictions with respect to the input image at each time step. We then computed the sensitivity of the value function prediction to each pixel:
\begin{align}
g_t^{w,h} & = \sqrt{ \sum_{c=1}^3 |\partial \hat{V}_t / \partial I^{w,h,c}_t|^2}. \nonumber   
\end{align}
We smoothed these sensitivity estimates using a 2 pixel-wide Gaussian filter:
\begin{align}
\hat{g}_t^{w,h} & = \text{GaussianFilter}(g_t^{w,h}, \sigma= \text{2 pixels}). \nonumber  
\end{align}
We then normalized this quantity based on its statistics across time and pixels by computing the 97th percentile:
\begin{align}
g_{97} & = \text{97th percentile of } \hat{g}_t^{w,h} \text{ over all } t, w, h. \nonumber  
\end{align}
Input images were then layered over a black image with an alpha channel that increased to 1 based on the sensitivity calculation. Specifically, we used an alpha channel value of:
\begin{equation}
\alpha_t^{w, h} = \min \bigg(0.3 + (1 - 0.3) \frac{\hat{g}_t^{w,h}}{g_{97}}, \, 1\bigg).
\end{equation}

\subsection{Learning Rate Analysis for High Discount Factor}
\label{sec:lranalysis}

To check that the learning rates used for the high discount RMA and LSTM models were reasonable, we ran the largest variance tasks from in Section~\ref{sec:imagevar} (for RMA with $\gamma=0.998$) and~\ref{sec:ktdvar} (for LSTM with $\gamma=0.998$) for learning rates $3.2 \times 10^{-7}$, $8 \times 10^{-7}$, $2 \times 10^{-6}$, $5 \times 10^{-6}$ and $1.25 \times 10^{-5}$. The results are shown in Figure~S\ref{fig:suppfig9} and they show that the default learning rate of $5 \times 10^{-6}$ was the best among those tried.

\subsection{Behavioral Analysis of Active Visual Match}
We compared the P1 behaviors of a TVT agent versus an RMA as shown in Figure 3a in the main text. First, we modified the environment to fix the color square in one of three pre-selected wall locations. We then ran TVT and RMA for 10 episodes in each of these three fixed color square conditions. Finally, we plotted the agents' positional trajectories in each condition. We also visualized the TVT agent's memory retrievals by plotting a single episode trajectory with arrowheads indicating agent orientation on each second agent step. Each arrowhead is also color-coded by the maximal read weight from any time step in P3 back to the memory encoded at this time and position in P1.

\subsection{Behavioral Analysis of Latent Information Acquisition}
We evaluated TVT and RMA for $50$ episodes in the latent information acquisition task. To visualize, we scatter-plotted the agent's position as a black dot for each P1 time step ($50$ episodes $\times$ $75$ P1 time steps = $3,750$ dots in total). We also binned the agent's position on a $4\times5$ grid and counted the percentage of time the agent had occupied each grid cell. We visualized this grid occupancy using a transparent heatmap overlying the top-down view.
To further quantify the behaviour of TVT versus RMA, we recorded how many objects were acquired by the agent in the exploration phase in each of the $50$ test trials and plotted the mean and standard deviation in a bar plot.

\subsection{Return Variance Analysis}
Over 20 trials, in Key-to-Door we computed and compared two return variances based on trajectories from the same TVT agent. The first was the undiscounted return: $R_t = \sum_{t' \geq t} r_{t'}$. The second was computed as in Algorithm~\ref{alg:rma} and Algorithm~\ref{alg:TVT} using TVT ($\gamma=\lambda=0.96$), i.e., it was bootstrapped recursively:
\begin{align*}
\tilde{R}_t = r_t + \gamma [\lambda \tilde{R}_{t+1} + (1-\lambda) \hat{V}_{t+1} ],  
\end{align*}
and $r_t$ was modified by TVT. 

\break
\clearpage

\section{Supplementary Figures}
\begin{figure}[!h]
    \centering
    \includegraphics[width=0.99\textwidth]{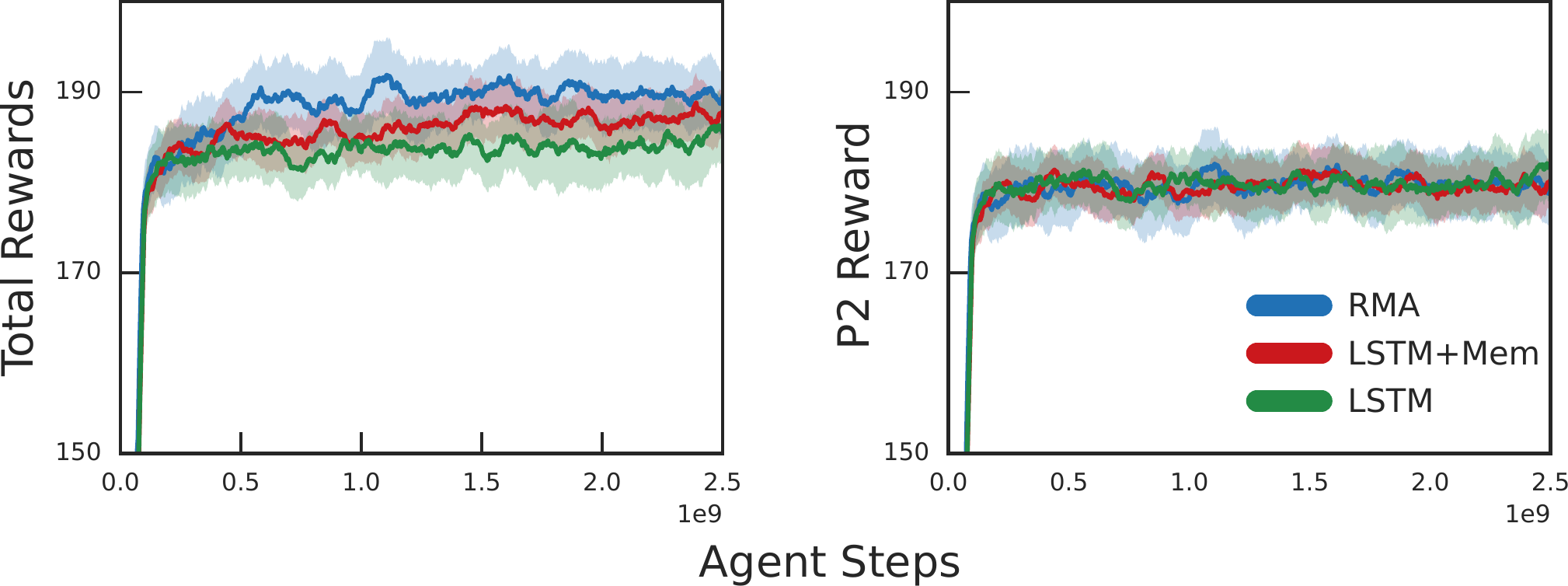}
    \caption{\textbf{Passive Image Match Learning.} \emph{Left.} Full episode score. \emph{Right.} P2 score. ($\gamma=0.96$ for all models.)}
    \label{fig:suppfig1}
\end{figure}

\begin{figure}[!h]
    \centering
    \includegraphics[width=0.99\textwidth]{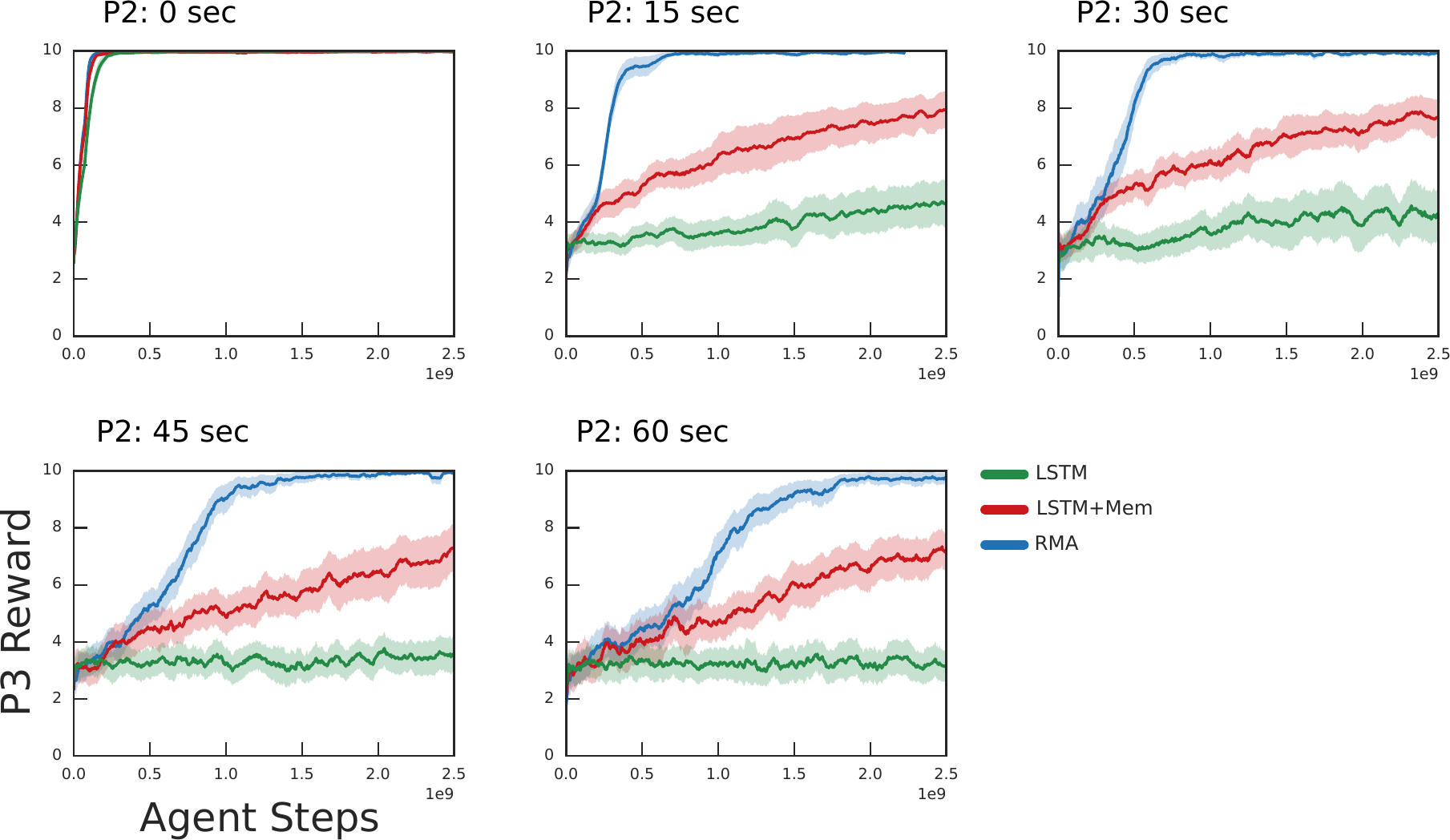}
    \caption{\textbf{Passive Image Match with Varying Delay Period.} All models learned to retrieve the P3 reward with no P2 delay, but performance is hampered for longer delays for models with no reconstructive loss.}
    \label{fig:suppfig2}
\end{figure}

\begin{figure}[!h]
    \centering
    \includegraphics[width=0.99\textwidth]{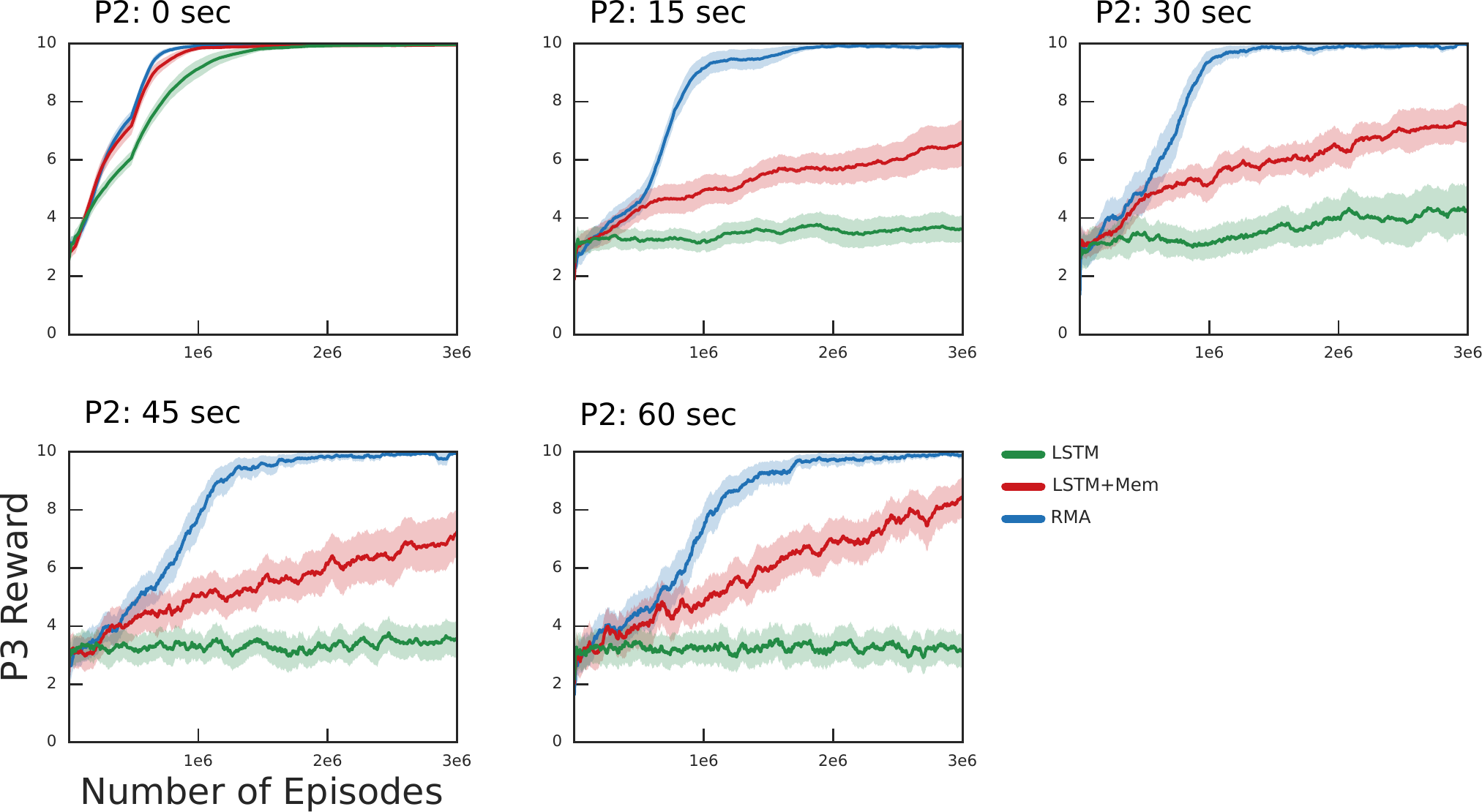}
    \caption{\textbf{Passive Image Match with Varying Delay Period (Episodes).} With the x-axis plotted in episodes, controlling for the number of additional steps due to the delay period, the RMA learned in roughly the same number of episodes, regardless of delay length (0 seconds to 60 seconds).}
    \label{fig:suppfig3}
\end{figure}

\begin{figure}[!h]
    \centering
    \includegraphics[width=0.99\textwidth]{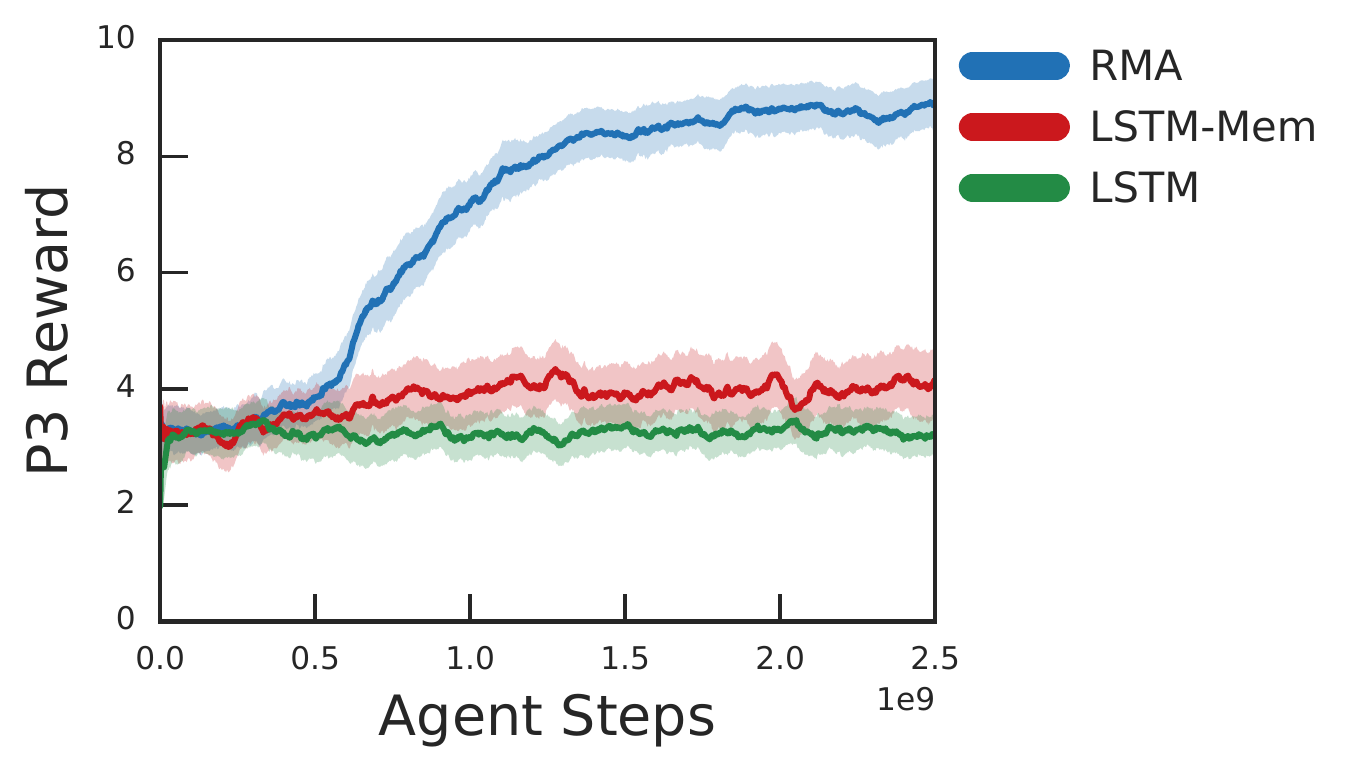}
    \caption{\textbf{Passive Image Match (CIFAR-10).} Using CIFAR-10 images\cite{krizhevsky2014cifar} instead of colored squares as P1 and P3 images, the RMA was still able to perform the Passive Image Match Task.}
    \label{fig:suppfig4}
\end{figure}

\begin{figure}[!h]
    \centering
    \includegraphics[width=0.99\textwidth]{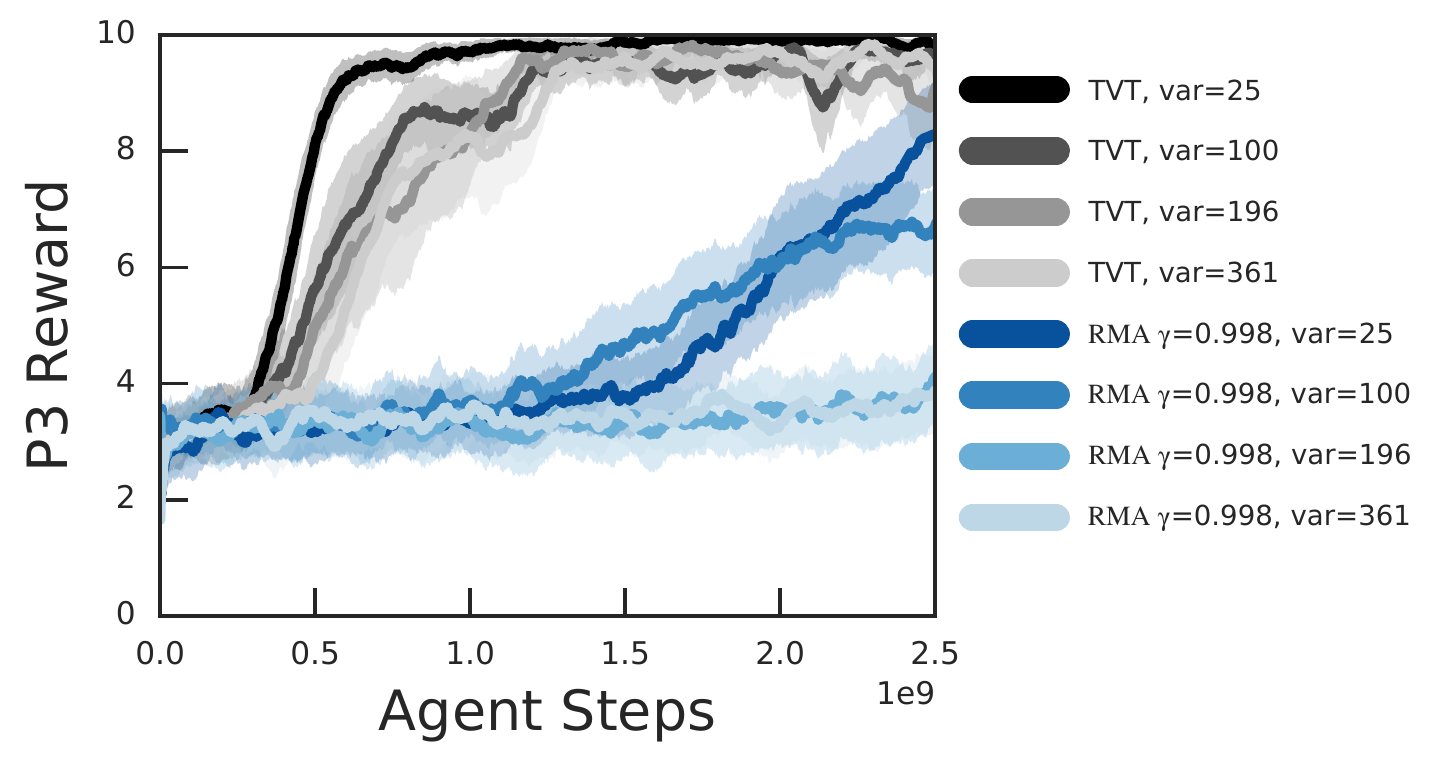}
    \caption{\textbf{Effect of P2 Reward Variance in Active Image Match.} P2 reward variance was introduced by varying the probability and reward value of apple reward (see \ref{sec:variableapple}). For higher levels of P2 reward variance, the RMA models failed to solve Active Image Match, though TVT was largely unaffected.}
    \label{fig:suppfig13}
\end{figure}

\begin{figure}[!h]
    \centering
    \includegraphics[width=0.99\textwidth]{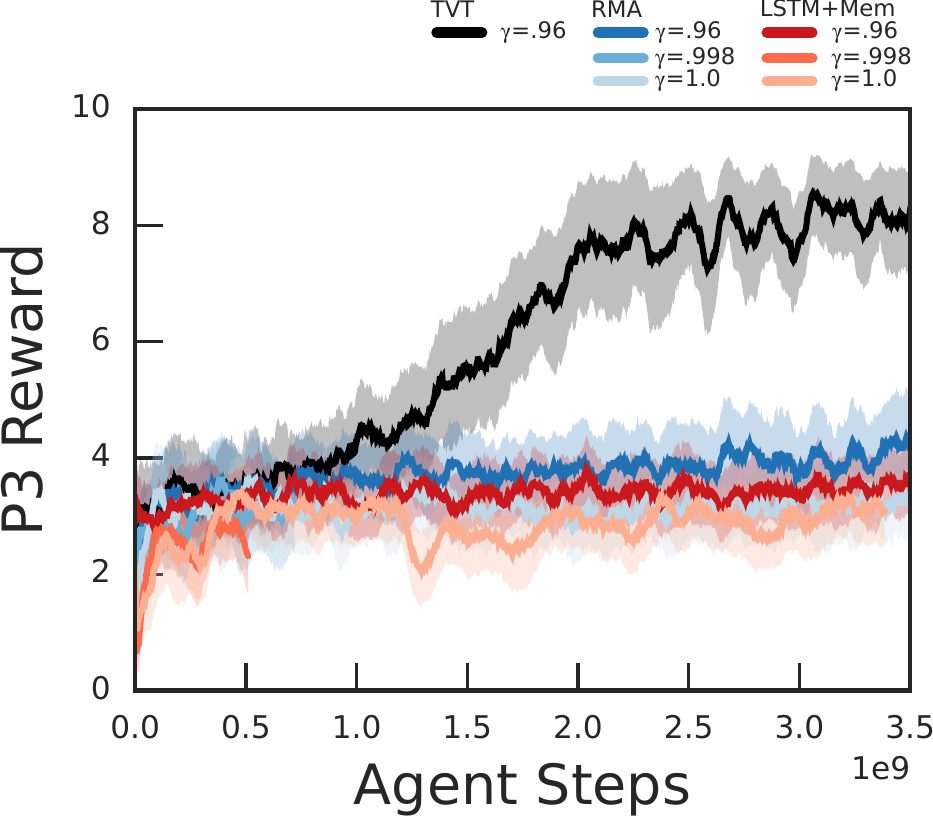}
    \caption{\textbf{Active Image Match 60 Second P2.} The TVT agent was also able to solve an Active Image Match task with a 60 second P2 delay period.}
    \label{fig:suppfig6}
\end{figure}

\begin{figure}[!h]
    \centering
    \includegraphics[width=0.99\textwidth]{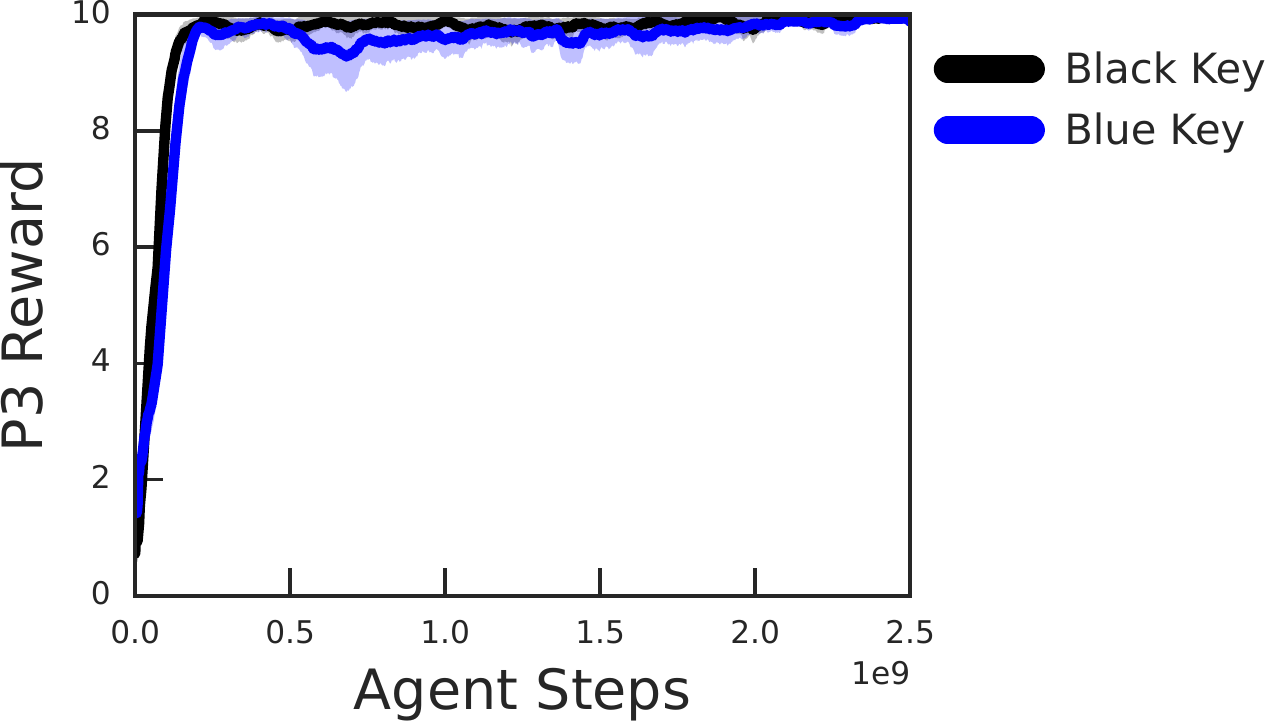}
    \caption{\textbf{Key-to-Door: Black vs. Blue key.} With a black door in P3, TVT was able to solve the task as easily with a blue key in P1, implying that content-based memory retrieval was flexible and not based on surface similarity between the key and door color.}
    \label{fig:suppfig7}
\end{figure}

\begin{figure}[!h]
    \centering
    \includegraphics[width=0.99\textwidth]{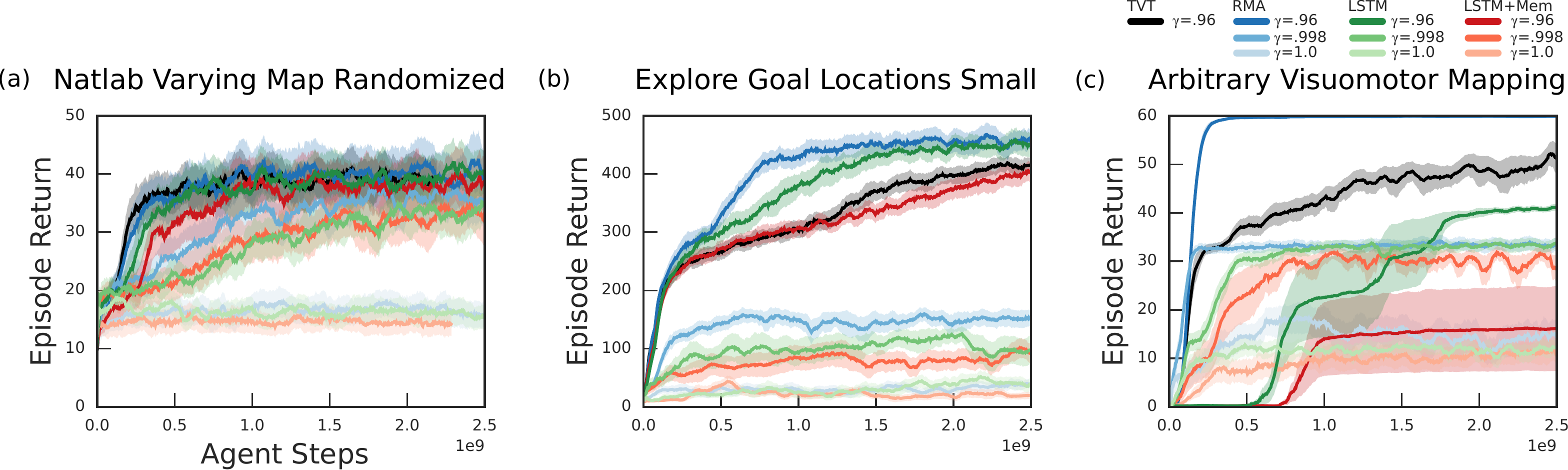}
    \caption{\textbf{Control Task DM Lab Learning.} \emph{a.} TVT (black) learned Natlab Varying Map Randomized just as well as the RMA. \emph{b.} On Explore Goal Locations Small, TVT led to a modest decrement in final performance. \emph{c.} On Psychlab Arbitrary Visuomotor Mapping, TVT did decrement final performance and slowed learning, though the agent's performance was still high compared to all but the RMA.}
    \label{fig:suppfig11}
\end{figure}

\begin{figure}[!h]
    \centering
    \includegraphics[width=0.99\textwidth]{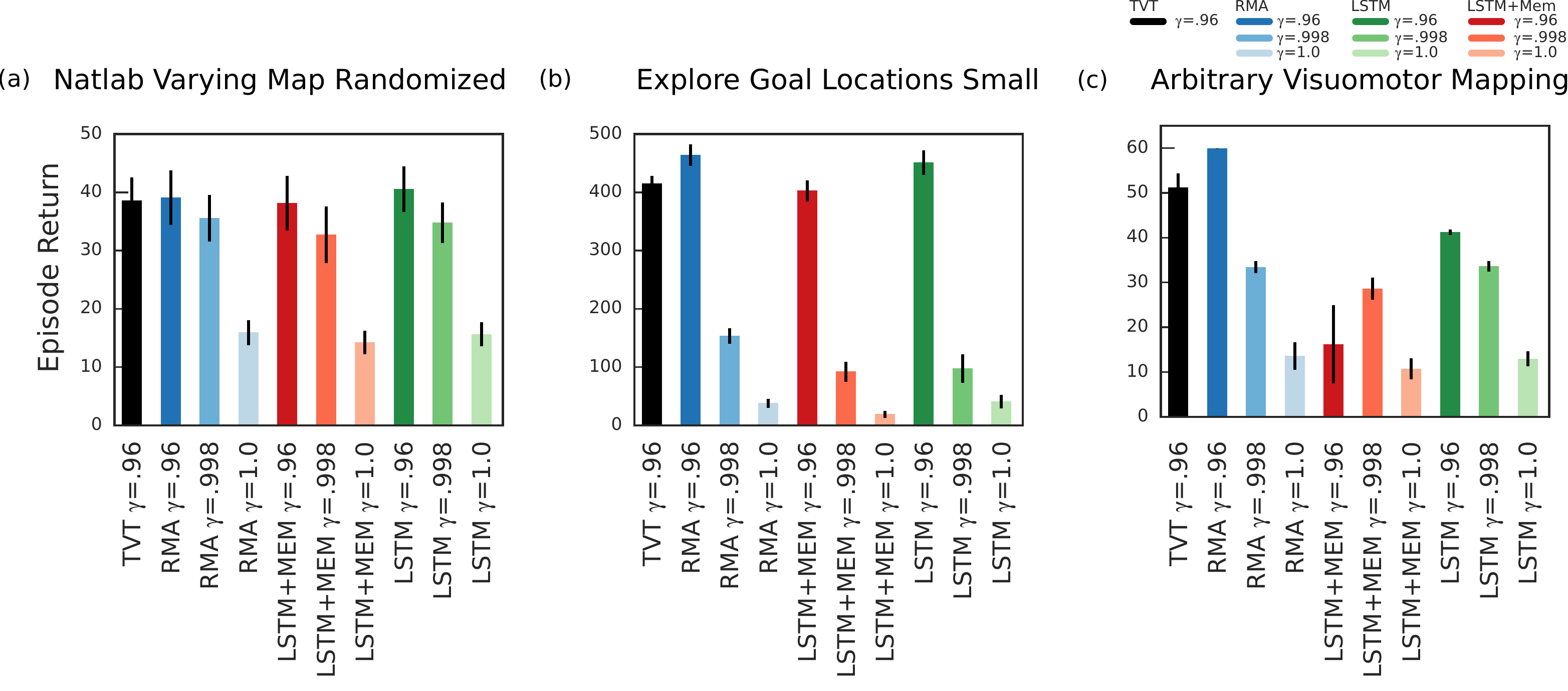}
    \caption{\textbf{Control Task DM Lab Final Performance.} Final performance for 5 training runs from Supplementary Figure~\ref{fig:suppfig11}.}
    \label{fig:suppfig12}
\end{figure}

\begin{figure}[!h]
    \centering
    \includegraphics[width=0.99\textwidth]{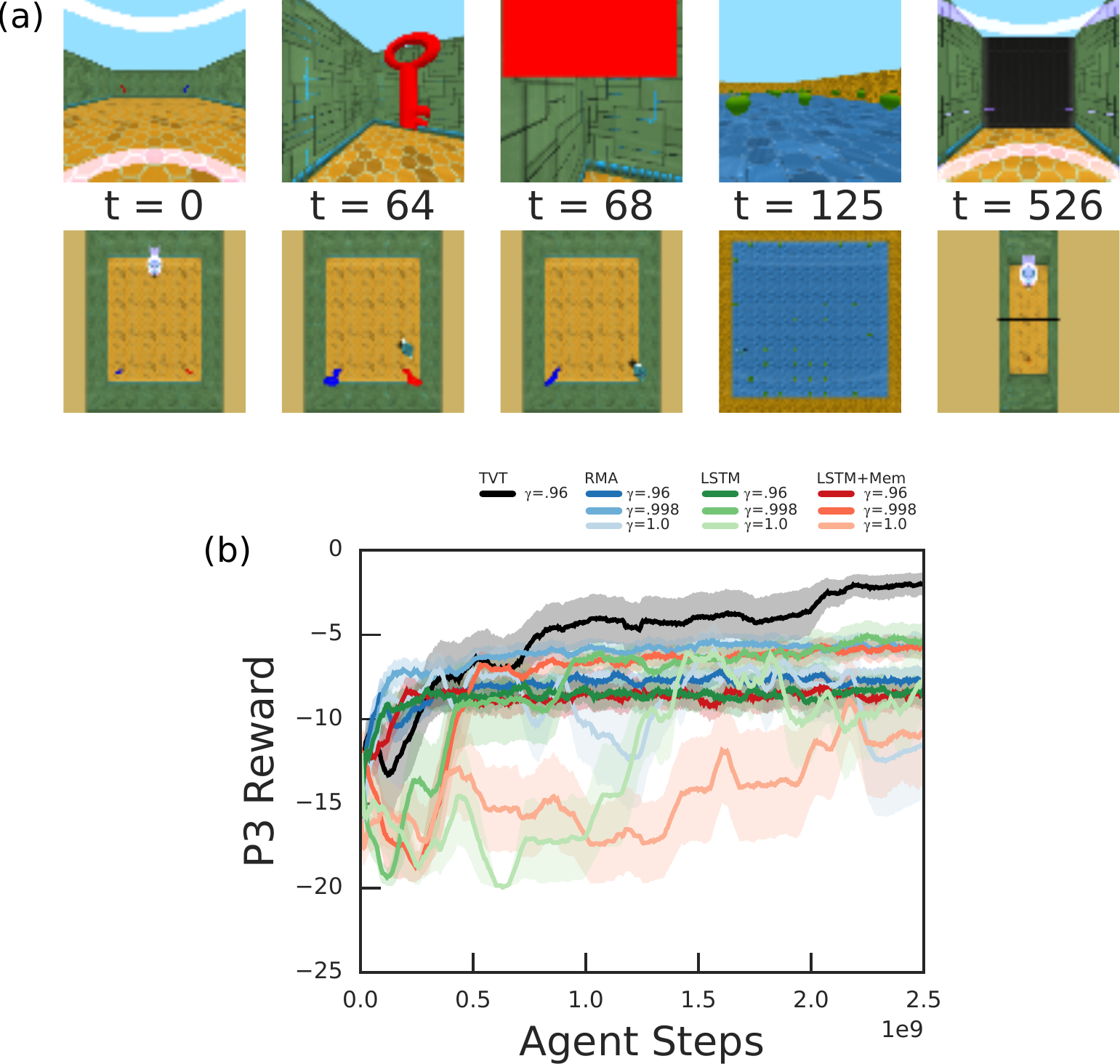}
    \caption{\textbf{Two Negative Keys level.} \emph{a.} In P1, the agent selects between a red and a blue key, distributed randomly in the room corners. The red key allows the agent to open the door in P3, receiving negative reward of $-1$. The blue key leads to negative reward of $-10$. No key selection leads to a negative reward of $-20$. \emph{b.} TVT was able to solve this task, picking up the red key, and receiving $-1$ on average in P3.}
    \label{fig:suppfig10}
\end{figure}

\begin{figure}[!h]
    \centering
    \includegraphics[width=0.99\textwidth]{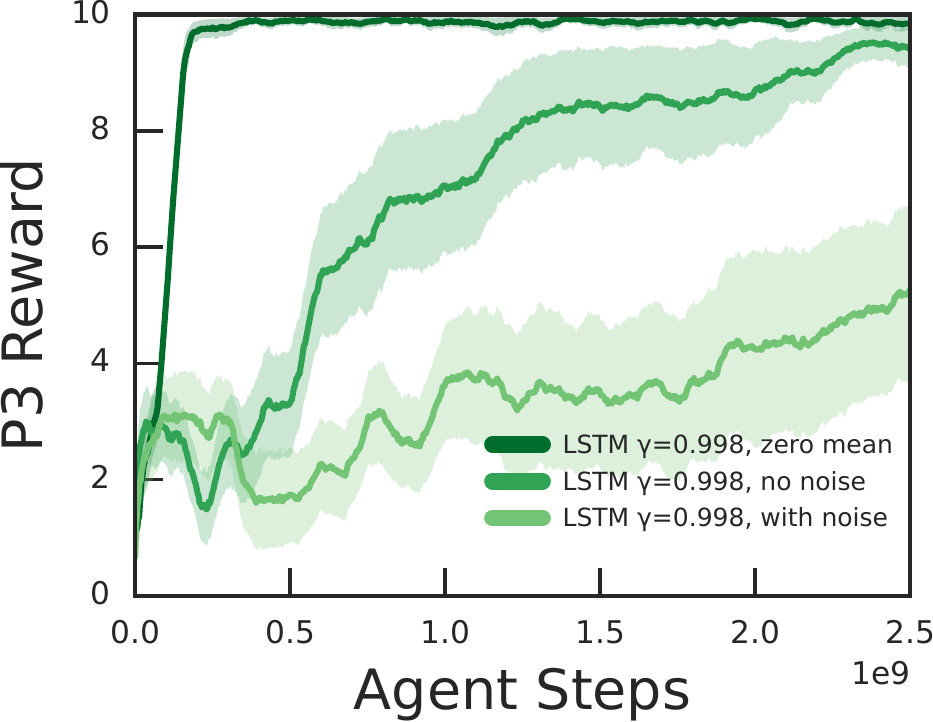}
    \caption{\textbf{Constant vs. Variable P2 Reward.} The three curves shown are for the LSTM agent with $\gamma=0.998$ in three variants of Key-to-Door: (i) zero apple reward (see~\ref{sec:zeroapple}), (ii) fixed number of apples each with reward 5 (see~\ref{sec:fixednumapples}), and (iii) the full level, which has a variable number of apples per episode but the same expected return as the fixed number of apples case. This analysis is discussed in Section~\ref{sec:ktdvar}. Variable P2 reward was maximally detrimental to performance.}
    \label{fig:suppfig8}
\end{figure}

\begin{figure}[!h]
    \centering
    \includegraphics[width=0.99\textwidth]{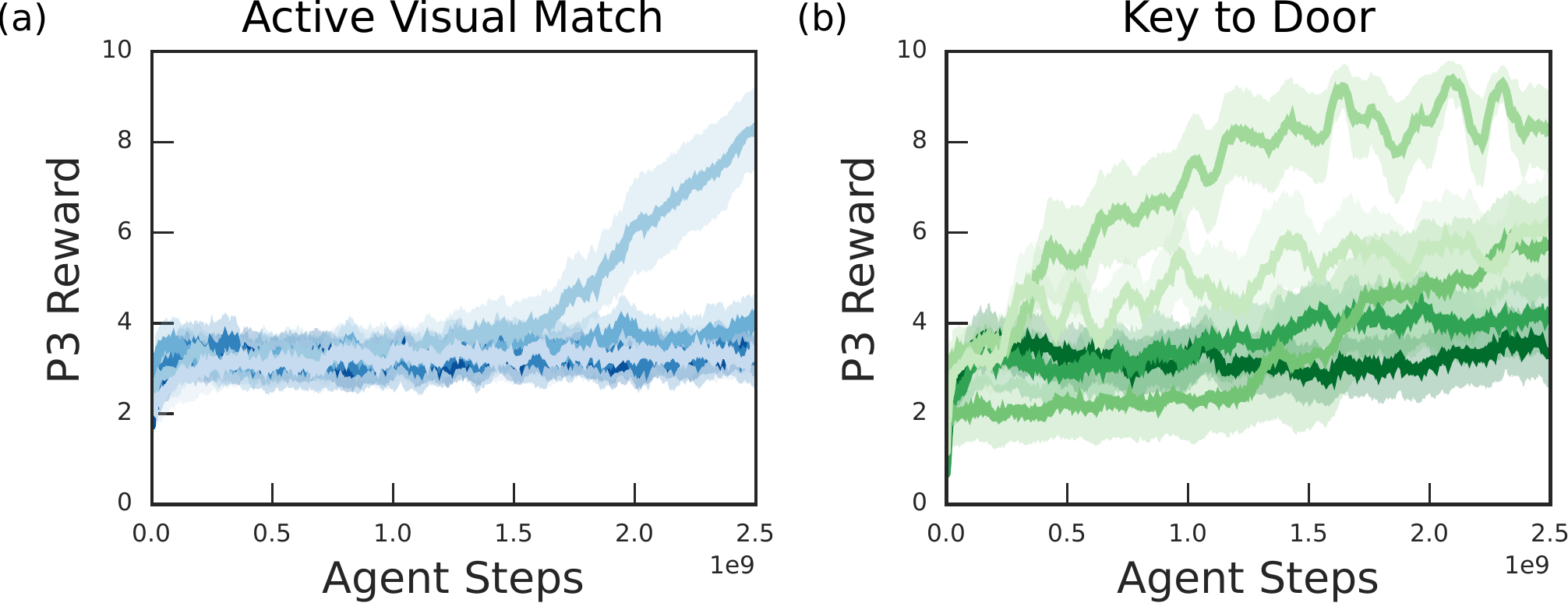}
    \caption{\textbf{Learning Rate Search on Comparison Models ($\gamma=0.998$).} Learning rates used were $3.2 \times 10^{-7}$, $8 \times 10^{-7}$, $2 \times 10^{-6}$, $5 \times 10^{-6}$, $1.25 \times 10^{-5}$, and are displayed from lightest to darkest in that order. In all analyses, the default learning rate of $5 \times 10^{-6}$ performed best. \emph{a.} RMA with $\gamma=0.998$ on Active Visual Match with apple reward $r_\text{ap[le}=1$. \emph{b.} LSTM with $\gamma=0.998$ on Key-to-Door task with variable apple reward as in Figure~4c in the main text, with P2 reward variance of 361.}
    \label{fig:suppfig9}
\end{figure}

\begin{figure}[!h]
    \centering
    \includegraphics[width=0.99\textwidth]{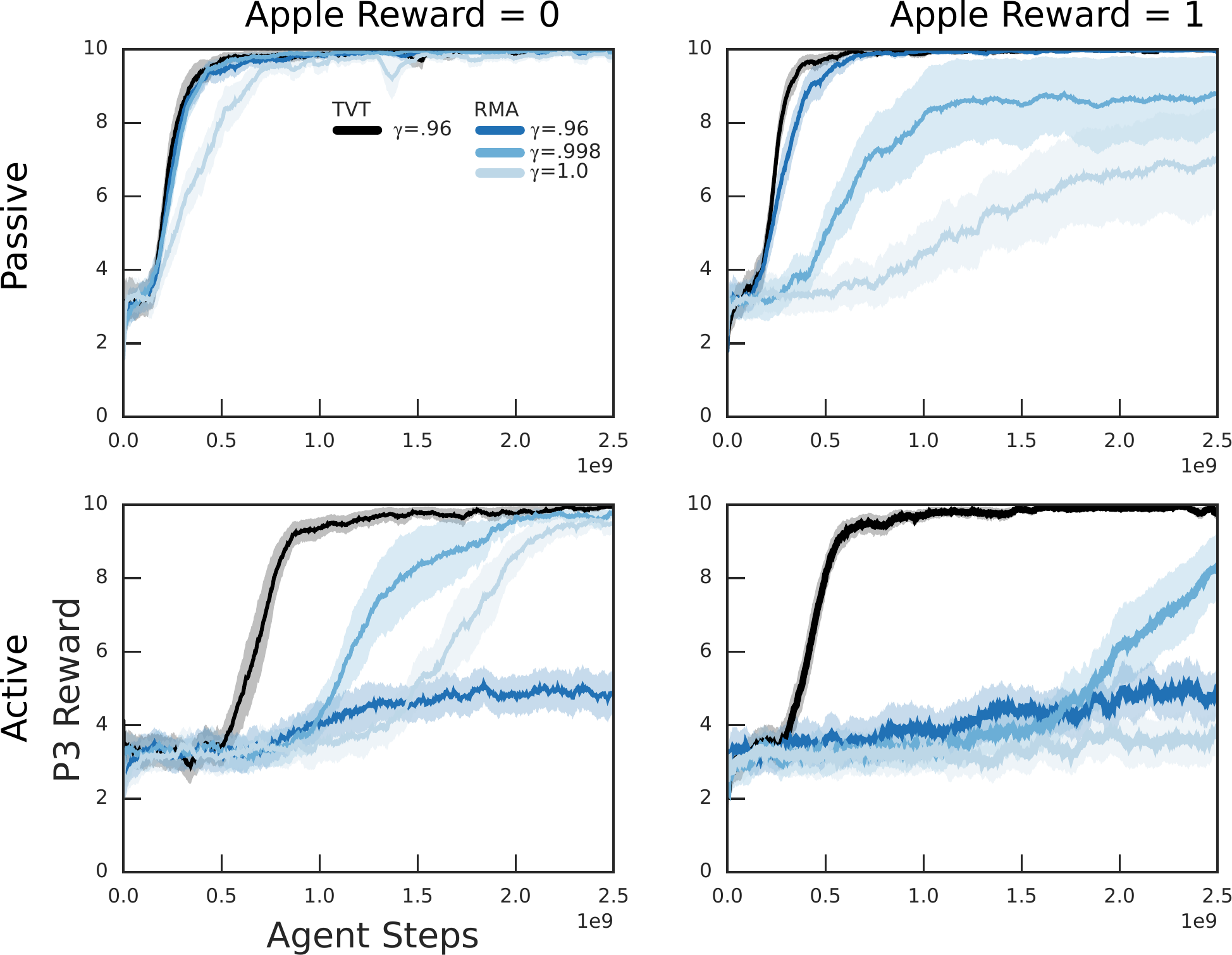}
    \caption{\textbf{Effect of P2 Apple Reward in Passive and Active Image Match Task.} \emph{Upper Row.} On Passive, the RMA performed worse with larger discount factors, which are not needed to solve the task. \emph{Lower Row.} On Active, the RMA models' performance at acquiring the distal reward degraded with the introduction of P2 reward. TVT remained stable with the introduction of P2 distractor reward.}
    \label{fig:suppfig5}
\end{figure}

\end{document}